\documentclass[letterpaper, 10 pt, conference]{ieeeconf}  

\IEEEoverridecommandlockouts                              

\usepackage{graphics} 
\usepackage{epsfig} 
\usepackage{mathptmx} 
\usepackage{times} 
\usepackage{amsmath} 

\usepackage{algorithm}
\usepackage{algpseudocode}
\usepackage{xcolor}
\usepackage{amssymb}
\usepackage{hyperref}

\usepackage{amsthm}

\newtheorem{remark}{Remark}
\newtheorem{definition}{Definition}
\newtheorem{lemma}{Lemma}
\newtheorem{proposition}{Proposition}

\usepackage{caption}
\usepackage{booktabs}

\usepackage{graphicx}
\usepackage{subcaption}

\usepackage{multirow}
\usepackage[most]{tcolorbox}

\title{\LARGE \bf
Efficient Transport-Driven Dynamic Safety Fields  \\ for Predictive Control
}

\author{Shanshan Zhao$^{1}$,
        Sheila Sutjipto$^{1}$,
        Lan Wu$^{2}$, and
        Teresa Vidal-Calleja$^{1}$%
\thanks{$^{1}$Shanshan Zhao, Sheila Sutjipto, and Teresa Vidal-Calleja
        are with the Robotics Institute, University of Technology Sydney,
        Sydney, NSW 2007, Australia.
        {\tt\small shanshan.zhao@student.uts.edu.au},
        {\tt\small \{sheila.sutjipto,teresa.vidalcalleja\}@uts.edu.au}.}%
\thanks{$^{2}$Lan Wu is with the School of Engineering,
        University of Western Australia,
        Perth, WA 6009, Australia.
        {\tt\small Lan.Wu@uwa.edu.au}.}%
}

\begin{document}

\maketitle
\thispagestyle{empty}
\pagestyle{empty}

\begin{abstract}
Safe motion in dynamic environments requires safety information that reflects the evolving environment. One such representation is a safety field, constructed by solving partial differential equations (PDEs), which provides a smooth and continuous representation of safety. However, as the environment changes, existing methods update the safety field by re-solving the PDEs from scratch, which is computationally expensive process that limits its applicability to real-time control. To address this challenge, we propose a method that efficiently updates the safety field by leveraging the transport equation combined with a harmonic correction. We further show that the updated field defines a safe set and retains a well-defined gradient, proving that it constitutes a valid discrete-time control barrier function (DCBF). Building on this, we integrate the safety field into an MPC framework, allowing the safety constraints over the prediction horizon to evolve with moving obstacles. Simulation and robotic manipulation experiments demonstrate that the proposed method reduces safety-field update time while supporting predictive safe control in dynamic environments.
\end{abstract}


\section{INTRODUCTION}
Safe robot motion in complex and time-varying environments demands that control objectives be achieved without compromising safety. 
Although motion planning and nominal control can generate task-oriented trajectories or control actions, they do not inherently provide explicit closed-loop safety guarantees in the presence of changing environmental constraints.
Control barrier functions (CBFs) provide a principled framework for safety-critical control \cite{ames2019control} where a scalar safety function defines a safe set and induces constraints on the control input to render this set forward invariant. CBF-based methods have consequently been widely adopted for safe robotic navigation and obstacle avoidance \cite{robot+cbf_1,robot+cbf_2,robot+cbf_3}. Thus, the construction of a suitable safety function that accurately represents the surrounding environment is a fundamental component of CBF-based robot control.

Signed Distance Fields (SDFs) are widely used to construct safety representations for collision avoidance and CBF-based control~\cite{xiao2021high,xiong2022discrete,wu2023log}. 
By assigning each point its signed distance to the nearest obstacle boundary, an SDF provides an intuitive and geometrically meaningful representation of free and occupied regions. 
However, because its value and gradient are primarily determined by the nearest boundary, an SDF does not explicitly account for the combined geometric influence of multiple surrounding boundaries.
This limitation becomes relevant in geometrically complex environments, such as those containing multiple obstacles, narrow passages, or confined workspaces.
The PDE-based approach in~\cite{Gilbert2025dynamic} constructs a safety field from boundary-normal information. 
By propagating boundary information throughout the workspace, the resulting field captures the combined geometric influence of surrounding boundaries while distinguishing safe and unsafe regions, thereby providing a suitable safety function for CBF-based control.

\begin{figure}[t]
    \centering
    \includegraphics[width=\columnwidth, trim=0 10 0 16, clip]{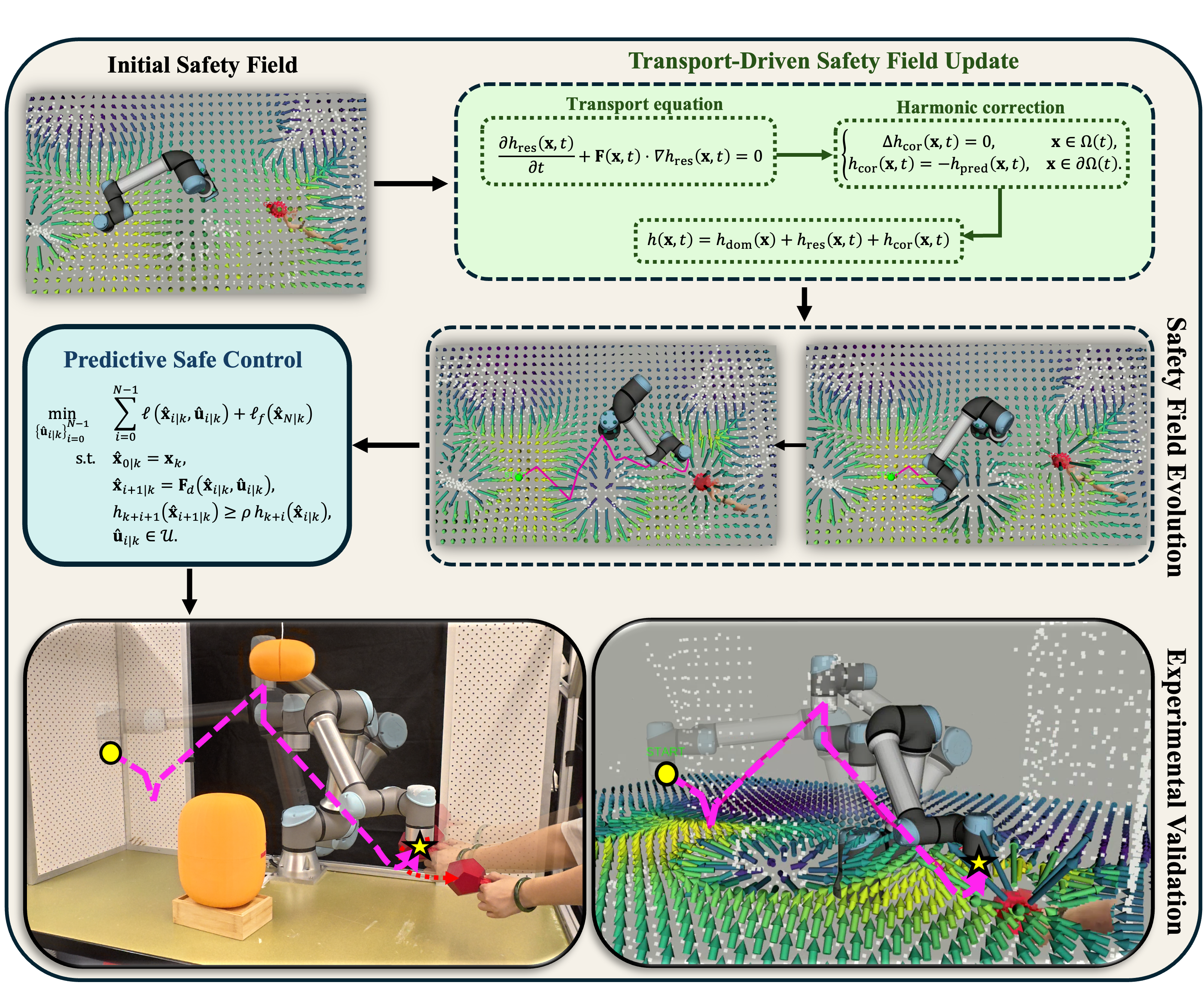}
    \caption{Overview of the proposed framework, where transport and correction efficiently update the safety field for predictive safe control in dynamic environments.}
    \label{fig:pipeline}
\end{figure}

Real-world robotic workspaces are rarely static, as pedestrians, mobile robots, manipulated objects, and other moving obstacles continuously reshape the surrounding free space.
As these obstacles move, the associated safety boundaries evolve over time, requiring the safety field to be updated accordingly. 
For PDE-based safety fields, a straightforward approach is to reconstruct the field at each time step by re-solving the underlying PDE with the updated boundary conditions \cite{bena2025geometry,yamaguchi2026layered,wilkinson2026full}. 
However, repeatedly solving the PDE over the entire workspace can introduce substantial computational cost, particularly when frequent field updates are required for real-time operation. 
This computational challenge motivates our development of an efficient safety-field update method for dynamic environments.

For safe control in dynamic environments, it is necessary to account not only for the current safety condition but also for how safety constraints may evolve as the robot moves. 
Model Predictive Control (MPC) provides a natural framework for this purpose by optimising control actions over a finite prediction horizon while explicitly considering system dynamics and constraints \cite{morari1999model}. 
In recent years, MPC has also been increasingly integrated with discrete-time control barrier functions (DCBFs) to enforce safety constraints over the prediction horizon, providing a predictive mechanism for safety-critical control.
Existing MPC-CBF approaches have mainly focused on embedding CBF constraints into the MPC prediction horizon \cite{zeng2021safety}, 
incorporating predicted obstacle motion for dynamic obstacle avoidance \cite{jian2023dynamic}, 
or combining MPC planning with lower-level CBF-based safety control \cite{rosolia2020multi}. 
In settings with dynamic obstacles, safety constraints are commonly constructed from explicit geometric representations of obstacles along the prediction horizon. 
In contrast, when safety is represented by a PDE-based field defined over the workspace, obstacle motion causes the field to evolve over time. 
Consequently, the safety constraint at each prediction step should be evaluated using the safety field corresponding to that future time. 
This motivates the integration of the proposed time-varying safety field with MPC through discrete-time CBF constraints.

The main contributions of this work are summarised as follows:
\begin{enumerate}
    \item We propose a transport-driven framework that efficiently updates PDE-based safety fields in dynamic environments through the transport equation and a harmonic correction, avoiding repeated full reconstruction of the safety field at every time step.
    
    \item We prove that the updated safety field satisfies the conditions of a valid DCBF, including a well-defined safe set and spatial gradient.

    \item We integrate the resulting time-varying safety field into an MPC framework with DCBF constraints, enabling predictive safe control under moving obstacles over the prediction horizon.
\end{enumerate}
Fig.~\ref{fig:pipeline}  illustrates the overall pipeline of our proposed approach.
The remainder of this paper is organised as follows.
Section \ref{sec:preliminary} introduces the problem formulation and necessary preliminaries.
Section \ref{sec:time-varying cbf} presents the baseline approach and the proposed transport-driven method for time-varying safety-field update.
Section \ref{sec:control} incorporates the discrete-time safety field into an MPC framework as a safety constraint.
Section \ref{sec:simulation} and \ref{sec:experiment} present simulation and experimental results.
Finally, Section \ref{sec:conclusion} concludes the paper and discusses future work.

\section{Preliminary}\label{sec:preliminary}
We consider a robotic system governed by the discrete-time dynamics
\begin{align}
\mathbf{x}_{k+1}
=
\mathbf{F}_d(\mathbf{x}_k,\mathbf{u}_k),
\end{align}
where $\mathbf{x}_k\in\mathbb{R}^n$ and
$\mathbf{u}_k\in\mathbb{R}^m$ denote the system state and control input at
time step $k$, respectively. A set
$\mathcal{C}\subset\mathbb{R}^n$ is forward invariant if
$\mathbf{x}_0\in\mathcal{C}$ implies
$\mathbf{x}_k\in\mathcal{C}$ for all $k\geq0$.
In safety-critical control, $\mathcal{C}$ represents the safe set, and
forward invariance ensures that a trajectory initialised in $\mathcal{C}$
remains within the safe set.

For discrete-time systems, this safety requirement can be enforced using a DCBF.
\begin{definition}[DCBF~\cite{agrawal2017discrete}]
\label{def2:dcbf}
Let
$\mathcal{C}
=
\left\{
\mathbf{x}\in\mathbb{R}^n
\mid
h(\mathbf{x})\geq0
\right\}$.
A function
$h:\mathbb{R}^n\rightarrow\mathbb{R}$
is a discrete-time control barrier function (DCBF) if there exists
$\rho\in[0,1]$ such that, for every
$\mathbf{x}_k\in\mathcal{C}$, there exists an admissible control input
$\mathbf{u}_k$ satisfying
\begin{align}
h\!\left(
\mathbf{F}_d(\mathbf{x}_k,\mathbf{u}_k)
\right)
\geq
\rho\,h(\mathbf{x}_k).
\label{eq:dcbf}
\end{align}
\end{definition}

The DCBF condition provides a sufficient condition for preserving the safe set under discrete-time control.

\begin{lemma}[Forward Invariance under DCBF~\cite{xiong2022discrete}]
\label{lem:dcbf_forward_invariance}
If $h$ is a discrete-time control barrier function and
$\mathbf{u}_k$ satisfies~\eqref{eq:dcbf} at every time step, then
$\mathcal{C}$ is forward invariant.
\end{lemma}

In this work, $h(\mathbf{x})$ is obtained from a PDE-based safety field over the workspace, following the static formulation in~\cite{Gilbert2025dynamic}.  
Its zero level set represents the safety boundary, with $h(\mathbf{x}) \geq 0$ corresponding to the safe region. 
If the environment contains dynamic obstacles, the safety field must evolve consistently with their motion. This motivates the use of the transport equation, which describes the propagation of a scalar field under a prescribed velocity field.

\begin{definition}[Transport equation~{\cite[Sec. 2.1]{evans2022partial}}]
\label{def:transport_definition}
Let $\mathbf{b}\in\mathbb{R}^n$ be a fixed vector. The transport equation
with constant coefficients is the first-order linear PDE
\begin{align}
\partial_t u + \mathbf{b}\cdot\nabla u = 0,
\label{eq:transport1}
\end{align}
where $u=u(\mathbf{x},t):\mathbb{R}^n\times[0,\infty)\rightarrow\mathbb{R}$ is the unknown, $\mathbf{x} \in \mathbb{R}^n$ denotes the spatial variable, $t \ge 0$ denotes time and
$\nabla u=(\partial_{x_1}u,\ldots,\partial_{x_n}u)$ denotes the gradient of
$u$ with respect to the spatial variables $\mathbf{x}$.
\end{definition}

\section{Time-Varying PDE-Based Safety Field}\label{sec:time-varying cbf}

For safety-critical control in dynamic environments, the CBF must evolve over time to reflect changes in the surrounding obstacles. 
Since the CBF in this work is constructed from a PDE-based safety field, the key challenge is to efficiently update the underlying safety field as the environment changes.
To formalise this update problem, we let $\Omega_{\mathrm{dom}}$ denote the workspace domain,
whose boundary includes the static environment and any static obstacle boundaries,
and let
$\Omega_{\mathrm{obs}}(t)$ denote the time-varying obstacle region.

The corresponding free space is then given by
$\Omega(t) = \Omega_{\mathrm{dom}} \setminus \Omega_{\mathrm{obs}}(t)$,
with boundary
$\partial\Omega(t) = \partial\Omega_{\mathrm{dom}} \cup \partial\Omega_{\mathrm{obs}}(t)$.
Here, $\partial\Omega_{\mathrm{obs}}(t)$ is represented as the zero level set of a scalar function $\phi(\mathbf{x},t)$, i.e., $\partial\Omega_{\mathrm{obs}}(t) = \{\mathbf{x} \mid \phi(\mathbf{x},t)=0\}$.
As the obstacle boundary evolves over time, the PDE-based safety field must be updated accordingly. The objective is therefore to efficiently construct the time-varying safety field $h(\mathbf{x},t)$ while ensuring that it remains a valid CBF. 
To this end, we first introduce a full PDE re-solving baseline and then present our proposed transport-and-correction method for efficient construction of the time-varying safety field.

\subsection{Baseline Method: Full PDE Re-solving}
A straightforward approach to constructing the time-varying safety field is to fully re-solve the underlying PDEs at each time step based on the updated obstacle geometry. 
The underlying PDE-based construction and its regularity properties follow the static formulation established in~\cite{Gilbert2025dynamic}. 
The evolution of the obstacle boundary is described by the transport equation
\begin{align}\label{eq:transport}
\frac{\partial \phi(\mathbf{x},t)}{\partial t}
+
\mathbf{F}(\mathbf{x},t)\cdot\nabla \phi(\mathbf{x},t)
=
0,
\end{align}
where $\phi(\mathbf{x},t)$ is the level-set function introduced above and $\mathbf{F}(\mathbf{x},t)$ denotes the prescribed obstacle motion field.
As $\phi(\mathbf{x},t)$ evolves according to this equation, its zero level set correspondingly updates the obstacle boundary $\partial\Omega_{\mathrm{obs}}(t)$, and hence the free space $\Omega(t)$ and its boundary $\partial\Omega(t)$.

Given the updated boundary $\partial\Omega(t)$, the time-varying safety field $h(\mathbf{x},t)$ is reconstructed using the same PDE-based formulation as in the static case. 
Specifically, $h(\mathbf{x},t)$ is obtained by solving the Poisson equation
\begin{align}\label{eq:poisson}
\begin{cases}
\Delta h(\mathbf{x},t)=f(\mathbf{x},t),
& \mathbf{x}\in\Omega(t),\\
h(\mathbf{x},t)=0,
& \mathbf{x}\in\partial\Omega(t),
\end{cases}
\end{align}
where the forcing function is defined as
$f(\mathbf{x},t) = \nabla\cdot\mathbf{v}(\mathbf{x},t)$,
and the guidance vector field $\mathbf{v}(\mathbf{x},t)$ is obtained by solving Laplace equation
\begin{align}\label{eq:laplace}
\begin{cases}
\Delta \mathbf{v}(\mathbf{x},t)=0,
& \mathbf{x}\in\Omega(t),\\
\mathbf{v}(\mathbf{x},t)=\mathbf{n}(\mathbf{x},t),
& \mathbf{x}\in\partial\Omega(t),
\end{cases}
\end{align}
with $\mathbf{n}(\mathbf{x},t)$ denoting the boundary normal field constructed from the updated boundary $\partial\Omega(t)$.
The complete full PDE re-solving procedure is summarised in Algorithm \ref{alg:baseline_resolve}.
This strategy has been used to update the safety field in dynamic environments~\cite{bena2025geometry, Gilbert2025dynamic}. Although it directly reconstructs the safety field from the updated boundary, it requires repeated solutions of both the Laplace and Poisson equations at each time step.
This repeated full-domain computation can become a major computational bottleneck when the safety field is used online within a control loop.

\begin{algorithm}[htbp]
\caption{Baseline: Full PDE Re-solving}
\label{alg:baseline_resolve}
\begin{algorithmic}[1]
\Require $\phi(\mathbf{x},t_0)$, obstacle motion field $\mathbf{F}(\mathbf{x},t)$
\Ensure Time-varying safety field $h(\mathbf{x},t_k)$

\For{$k=1,2,\ldots$}
    \State Update obstacle boundary using~\eqref{eq:transport}
    \State Compute boundary normals $\mathbf{n}(\mathbf{x},t_k)$
    \State Obtain the guidance vector field $\mathbf{v}(\mathbf{x},t_k)$ by solving~\eqref{eq:laplace}
    \State Compute the forcing function $f(\mathbf{x},t_k)=\nabla\cdot\mathbf{v}(\mathbf{x},t_k)$
    \State Obtain the safety field $h(\mathbf{x},t_k)$ by solving~\eqref{eq:poisson}
\EndFor
\end{algorithmic}
\end{algorithm}

\subsection{Proposed Method: Transport-Driven Safety Field Update with Correction}

To avoid reconstructing the complete safety field from scratch at every time step, we represent the proposed time-varying safety field as
\begin{align}
h(\mathbf{x},t)
=
h_{\mathrm{dom}}(\mathbf{x})
+
h_{\mathrm{res}}(\mathbf{x},t)
+
h_{\mathrm{cor}}(\mathbf{x},t),
\end{align}
where $h_{\mathrm{dom}}(\mathbf{x})$ denotes the static field associated with the fixed domain boundary, $h_{\mathrm{res}}(\mathbf{x},t)$ denotes the obstacle-related residual field, and $h_{\mathrm{cor}}(\mathbf{x},t)$ is a harmonic correction term that enforces the boundary condition on the updated boundary $\partial\Omega(t)$.
For $N_{\mathrm{obs}}$ moving obstacles, the total residual field is defined as
\begin{align}
h_{\mathrm{res}}(\mathbf{x},t)
=
\sum_{i=1}^{N_{\mathrm{obs}}}
h_{\mathrm{res}}^{(i)}(\mathbf{x},t),
\end{align}
where $h_{\mathrm{res}}^{(i)}(\mathbf{x},t)$ denotes the residual field associated with the $i$th obstacle.

For each obstacle $i$, let $h^{(i)}(\mathbf{x},t_0)$ denote the safety field constructed for the fixed domain boundary and the $i$th obstacle only. 
The corresponding initial residual field is then defined as
\begin{align}
h_{\mathrm{res}}^{(i)}(\mathbf{x},t_0)
=
h^{(i)}(\mathbf{x},t_0) - h_{\mathrm{dom}}(\mathbf{x}),
\end{align}

\begin{figure*}[htbp]
\centering
\rule{0pt}{12pt}\par
\footnotesize

\noindent
\begin{minipage}[t]{0.49\textwidth}
\centering
\setlength{\tabcolsep}{1pt}
\begin{tabular}{@{}ccc@{}}
\makebox[0.31\linewidth][c]{$t=0\,\mathrm{s}$} &
\makebox[0.31\linewidth][c]{$t=2\,\mathrm{s}$} &
\makebox[0.31\linewidth][c]{$t=5\,\mathrm{s}$}
\end{tabular}
\end{minipage}%
\hfill%
\begin{minipage}[t]{0.49\textwidth}
\centering
\setlength{\tabcolsep}{1pt}
\begin{tabular}{@{}ccc@{}}
\makebox[0.31\linewidth][c]{$t=0\,\mathrm{s}$} &
\makebox[0.31\linewidth][c]{$t=2\,\mathrm{s}$} &
\makebox[0.31\linewidth][c]{$t=5\,\mathrm{s}$}
\end{tabular}
\end{minipage}

\vspace{2pt}

\begin{tcolorbox}[
    width=\textwidth,
    colframe=red,
    colback=white,
    boxrule=1.1pt,
    arc=1pt,
    left=2pt,
    right=2pt,
    top=2pt,
    bottom=2pt,
    boxsep=0pt
]

\begin{minipage}[t]{0.49\linewidth}
\centering

\begin{tcolorbox}[
    enhanced,
    width=\linewidth,
    colback=white,
    colframe=white,   
    boxrule=0pt,
    arc=1pt,
    left=1pt,
    right=1pt,
    top=1pt,
    bottom=1pt,
    boxsep=0pt,
    overlay={
        \draw[
            red!55,
            line width=0.6pt,
            dash pattern=on 3pt off 2pt
        ]
        (frame.south west) rectangle (frame.north east);
    }
]

\centering
\setlength{\tabcolsep}{1pt}

\begin{tabular}{@{}ccc@{}}

\includegraphics[width=0.31\linewidth]
{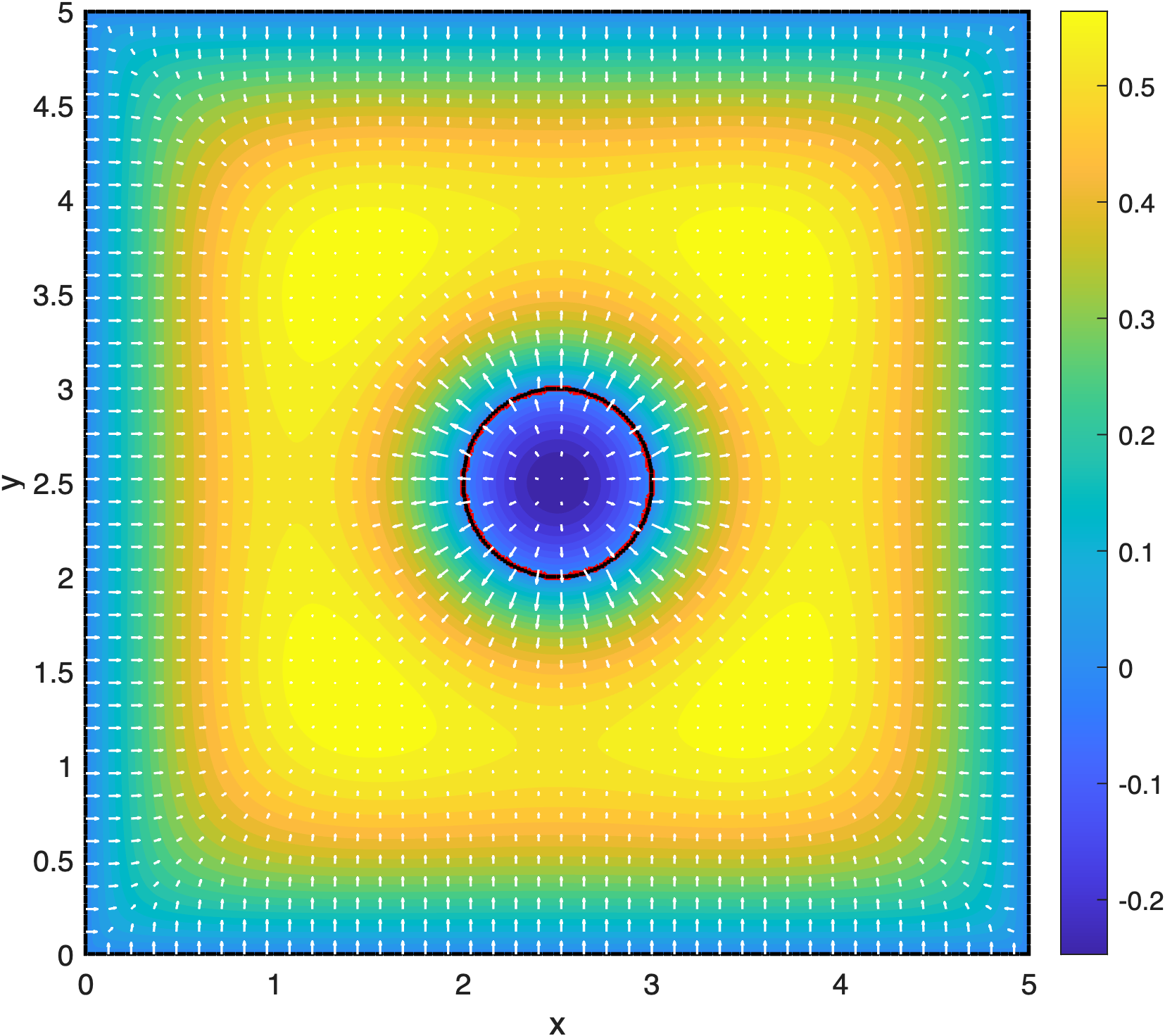}
&
\includegraphics[width=0.31\linewidth]
{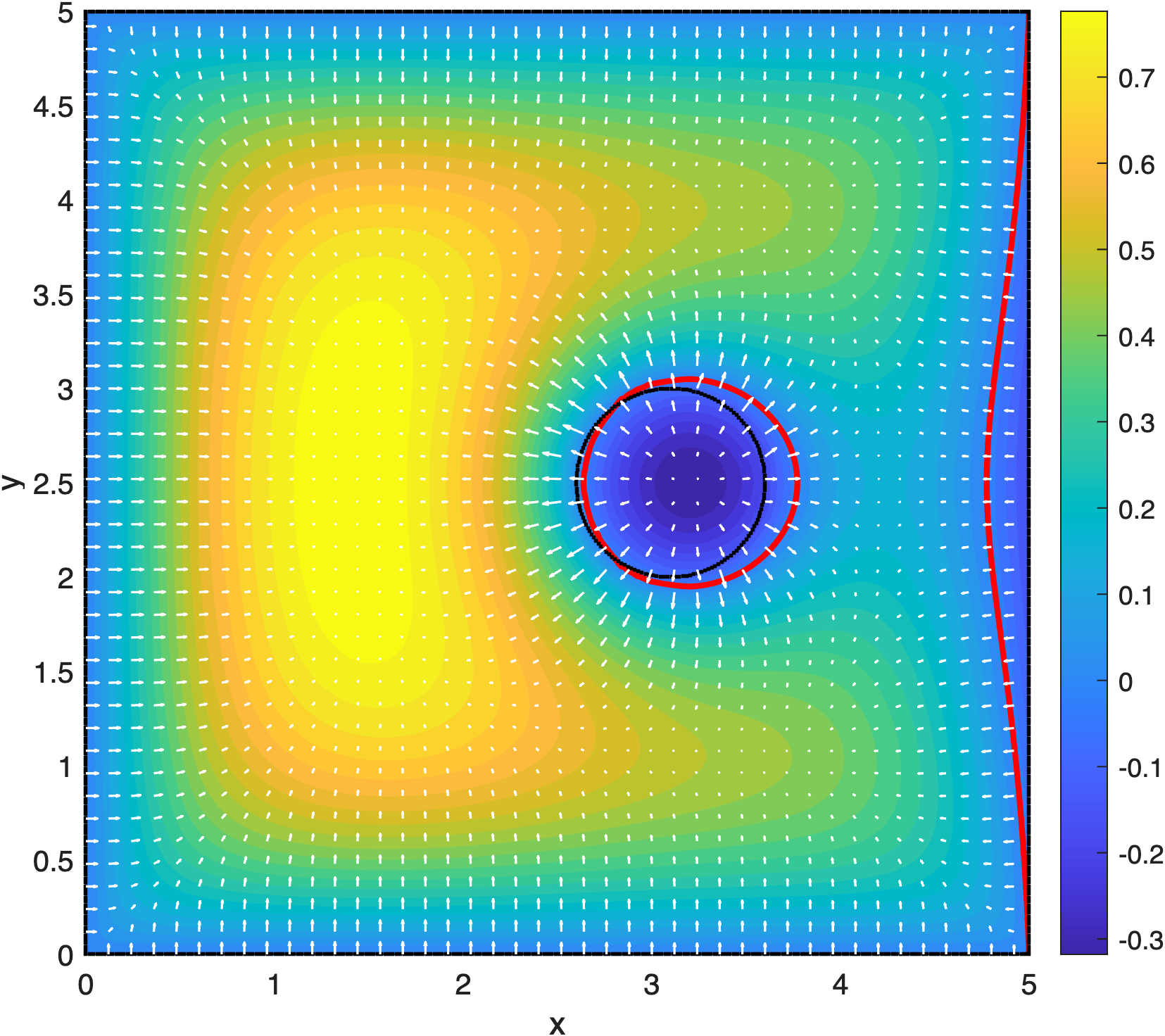}
&
\includegraphics[width=0.31\linewidth]
{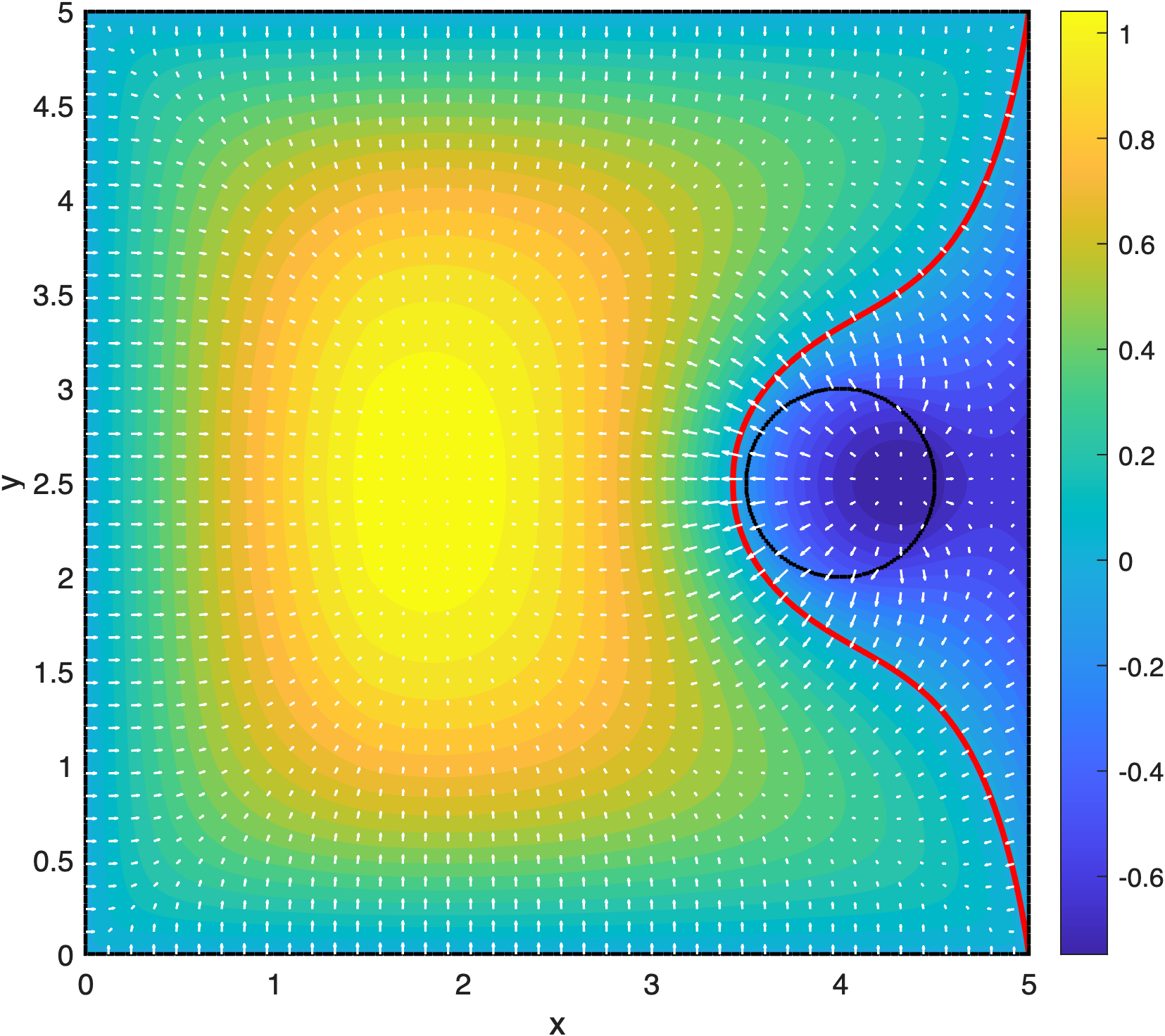}
\\[-1pt]

{\scriptsize (a)} &
{\scriptsize (b)} &
{\scriptsize (c)}
\\[2pt]

\includegraphics[width=0.31\linewidth]
{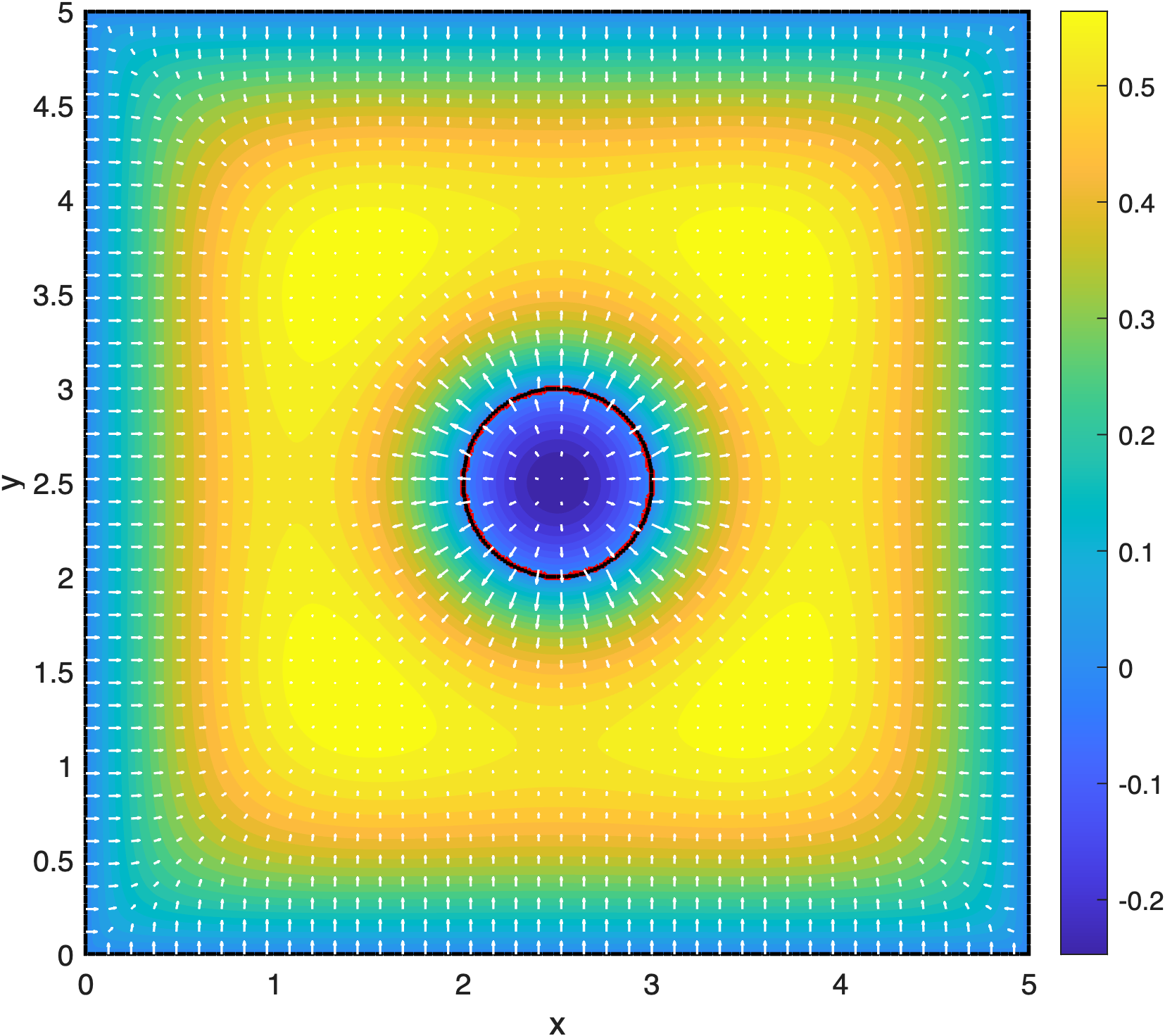}
&
\includegraphics[width=0.31\linewidth]
{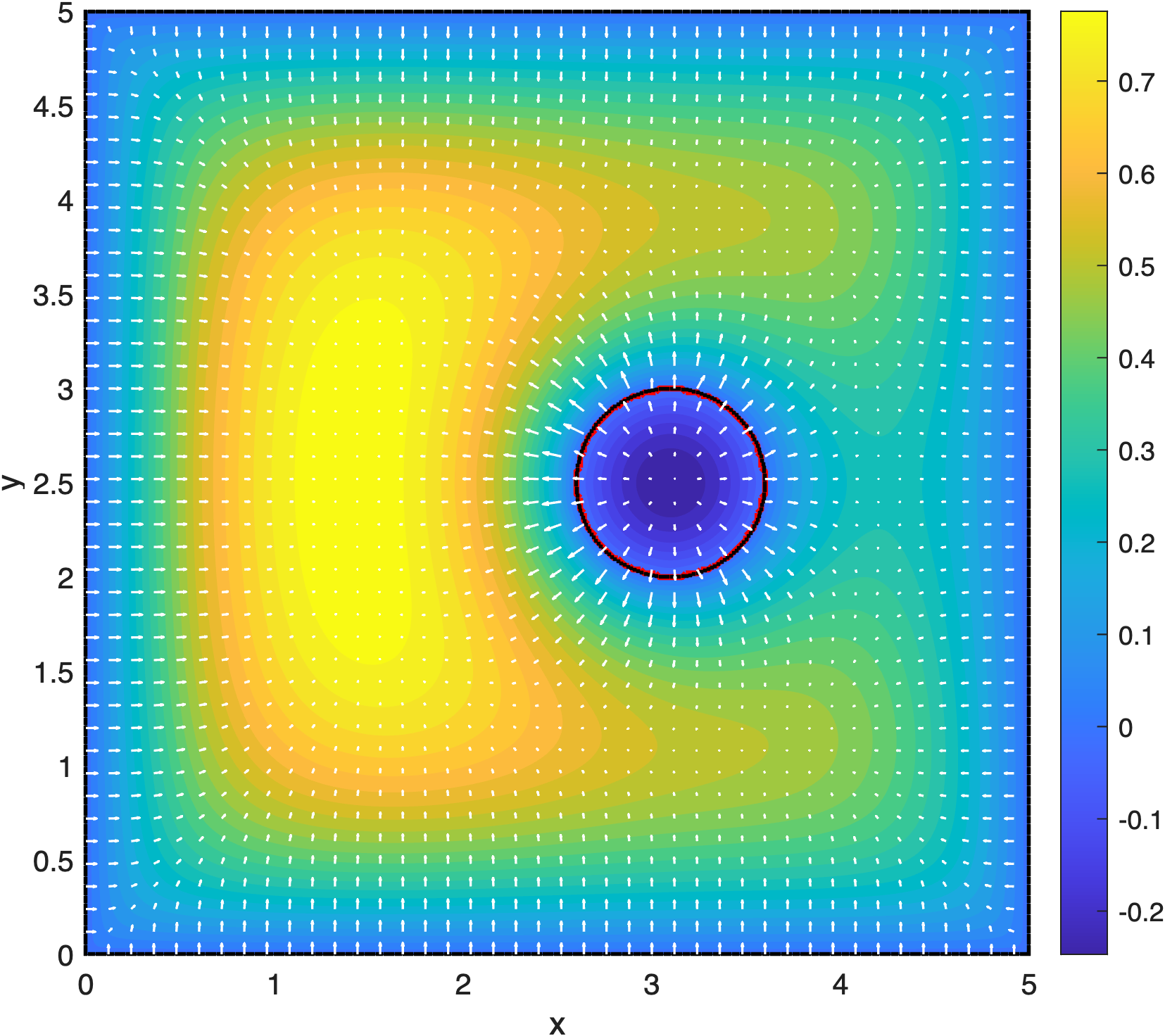}
&
\includegraphics[width=0.31\linewidth]
{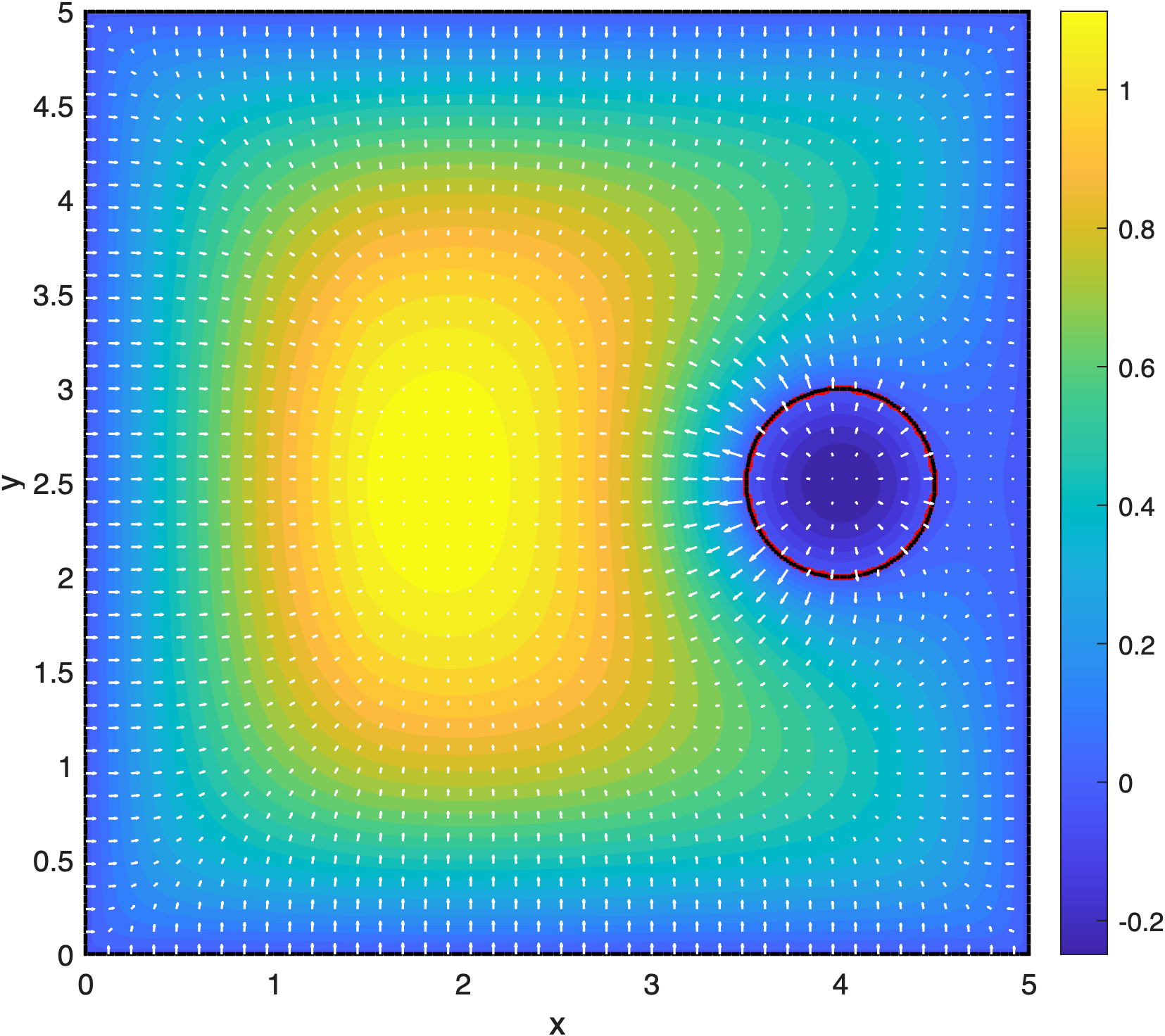}
\\[-1pt]

{\scriptsize (d)} &
{\scriptsize (e)} &
{\scriptsize (f)}

\end{tabular}

\end{tcolorbox}

\end{minipage}%
\hfill%
\begin{minipage}[t]{0.49\linewidth}
\centering


\begin{tcolorbox}[
    enhanced,
    width=\linewidth,
    colback=white,
    colframe=white,   
    boxrule=0pt,
    arc=1pt,
    left=1pt,
    right=1pt,
    top=1pt,
    bottom=1pt,
    boxsep=0pt,
    overlay={
        \draw[
            red!55,
            line width=0.6pt,
            dash pattern=on 3pt off 2pt
        ]
        (frame.south west) rectangle (frame.north east);
    }
]

\centering
\setlength{\tabcolsep}{1pt}

\begin{tabular}{@{}ccc@{}}

\includegraphics[width=0.31\linewidth]
{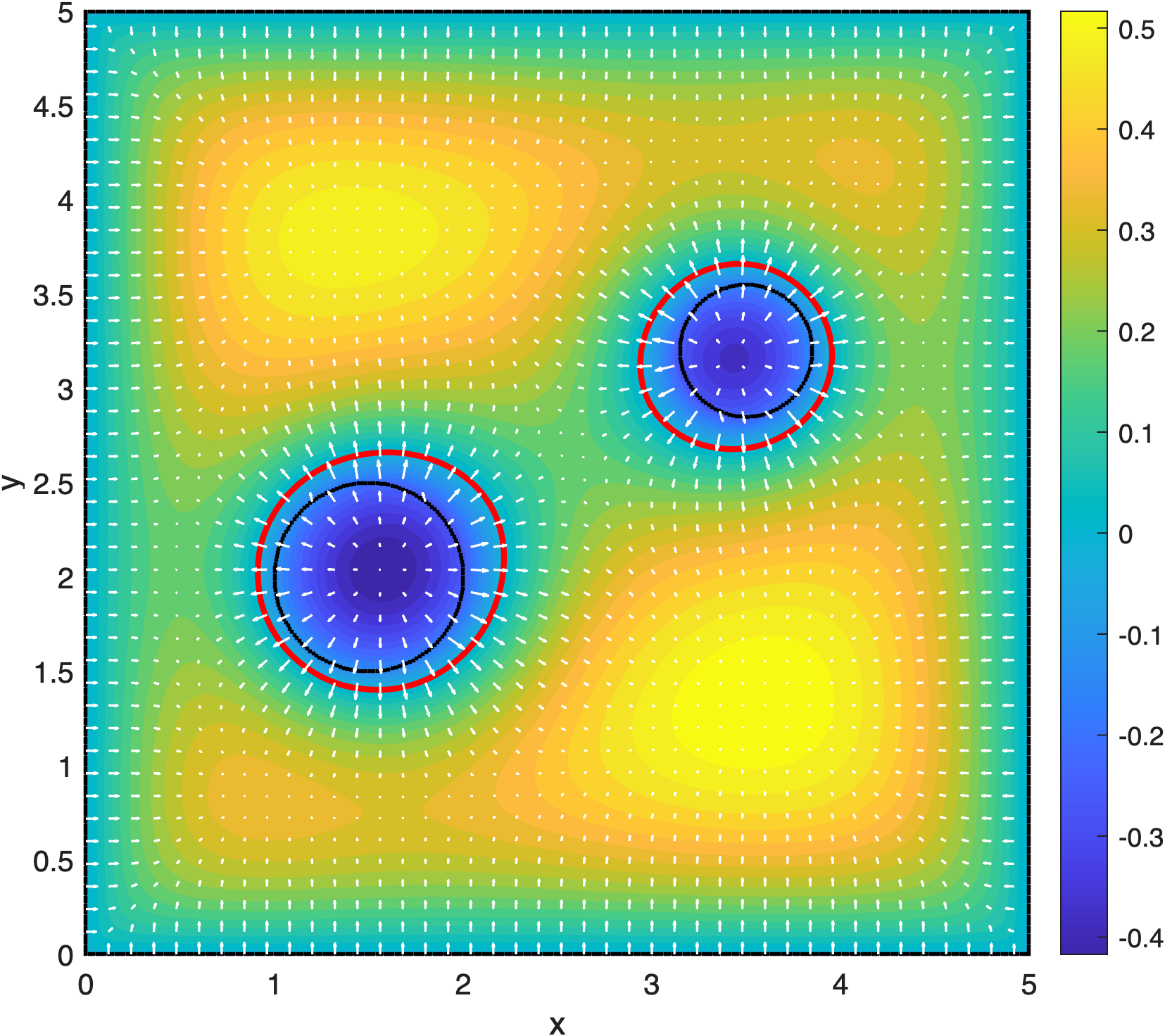}
&
\includegraphics[width=0.31\linewidth]
{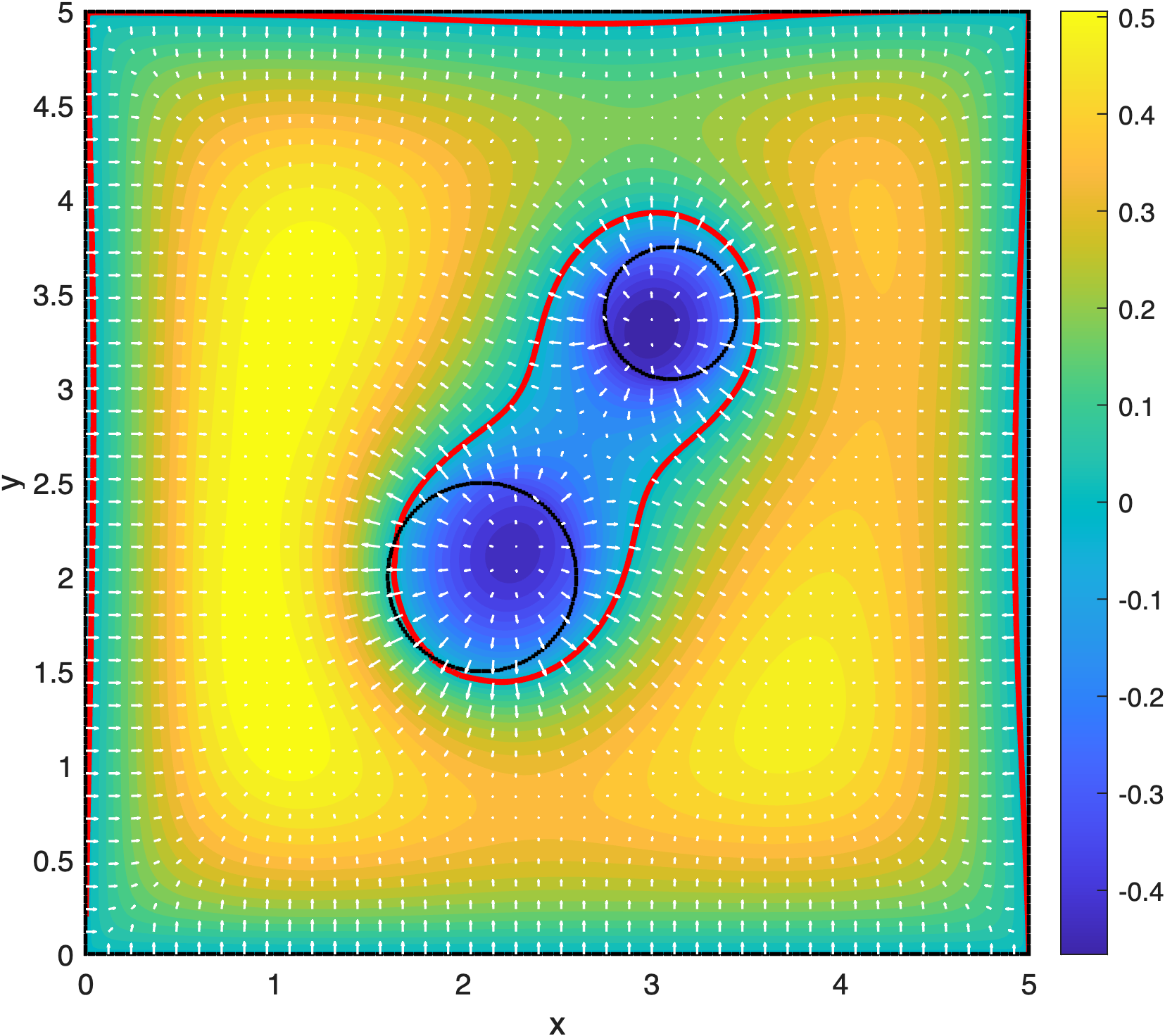}
&
\includegraphics[width=0.31\linewidth]
{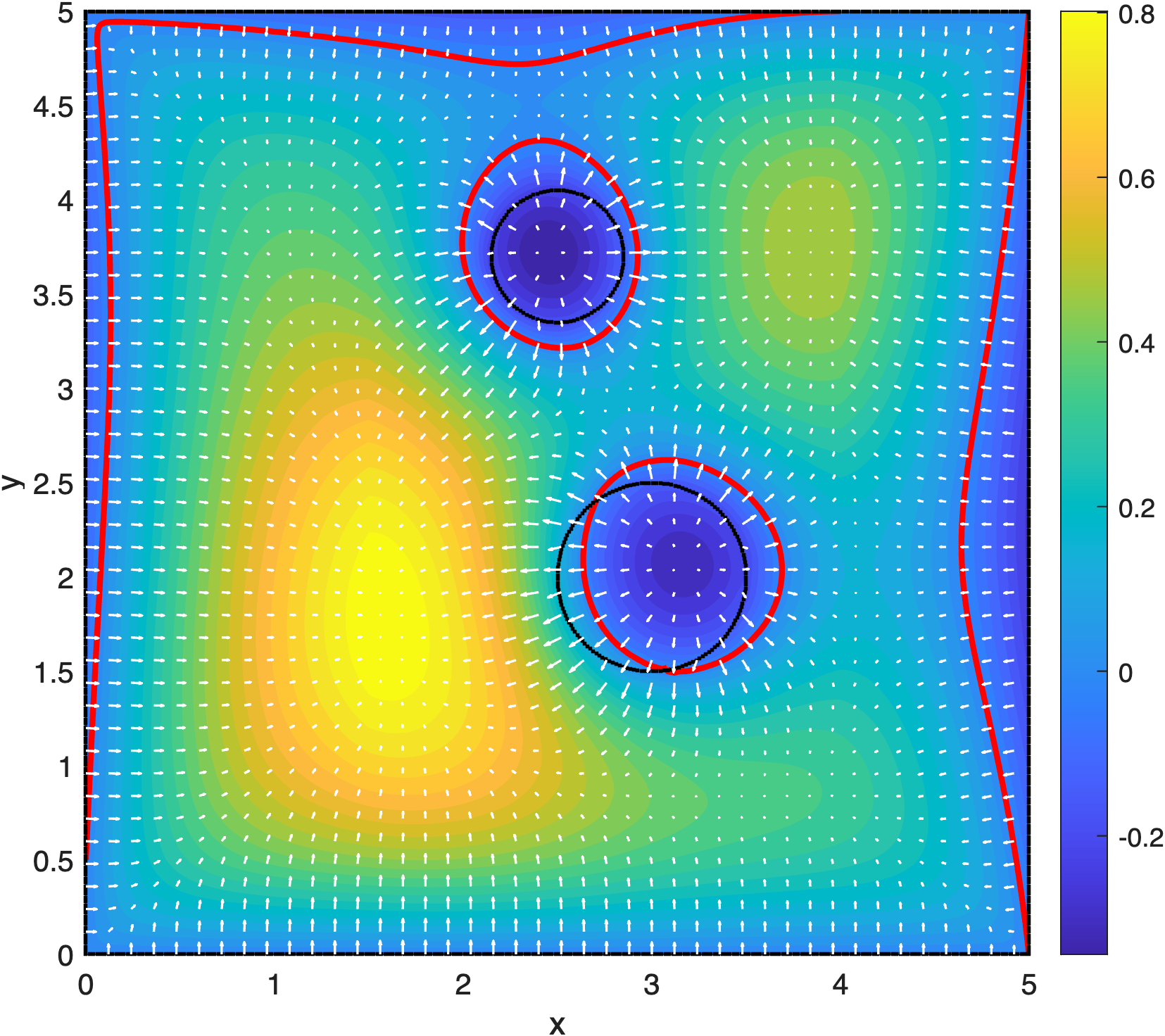}
\\[-1pt]

{\scriptsize (g)} &
{\scriptsize (h)} &
{\scriptsize (i)}
\\[2pt]

\includegraphics[width=0.31\linewidth]
{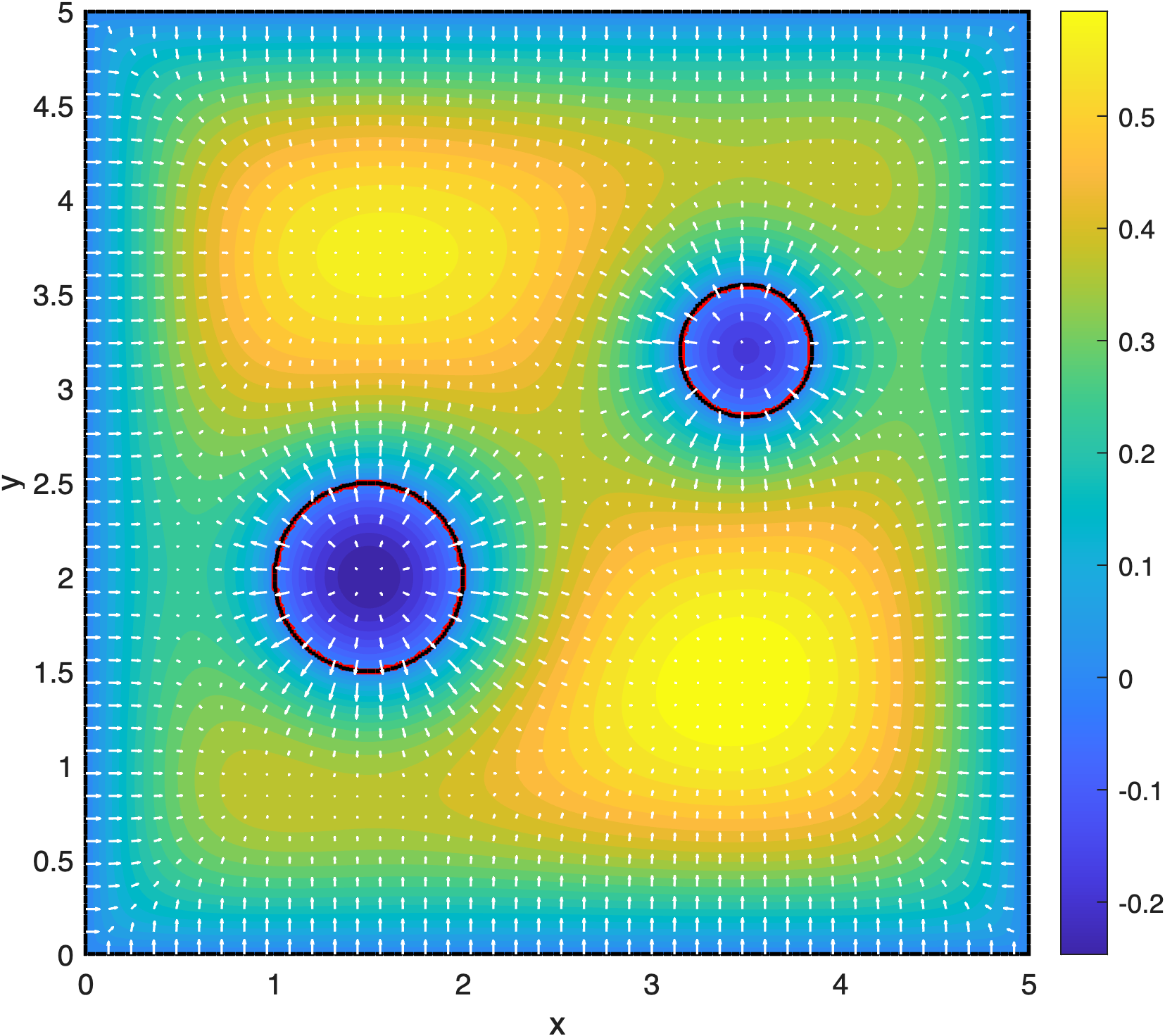}
&
\includegraphics[width=0.31\linewidth]
{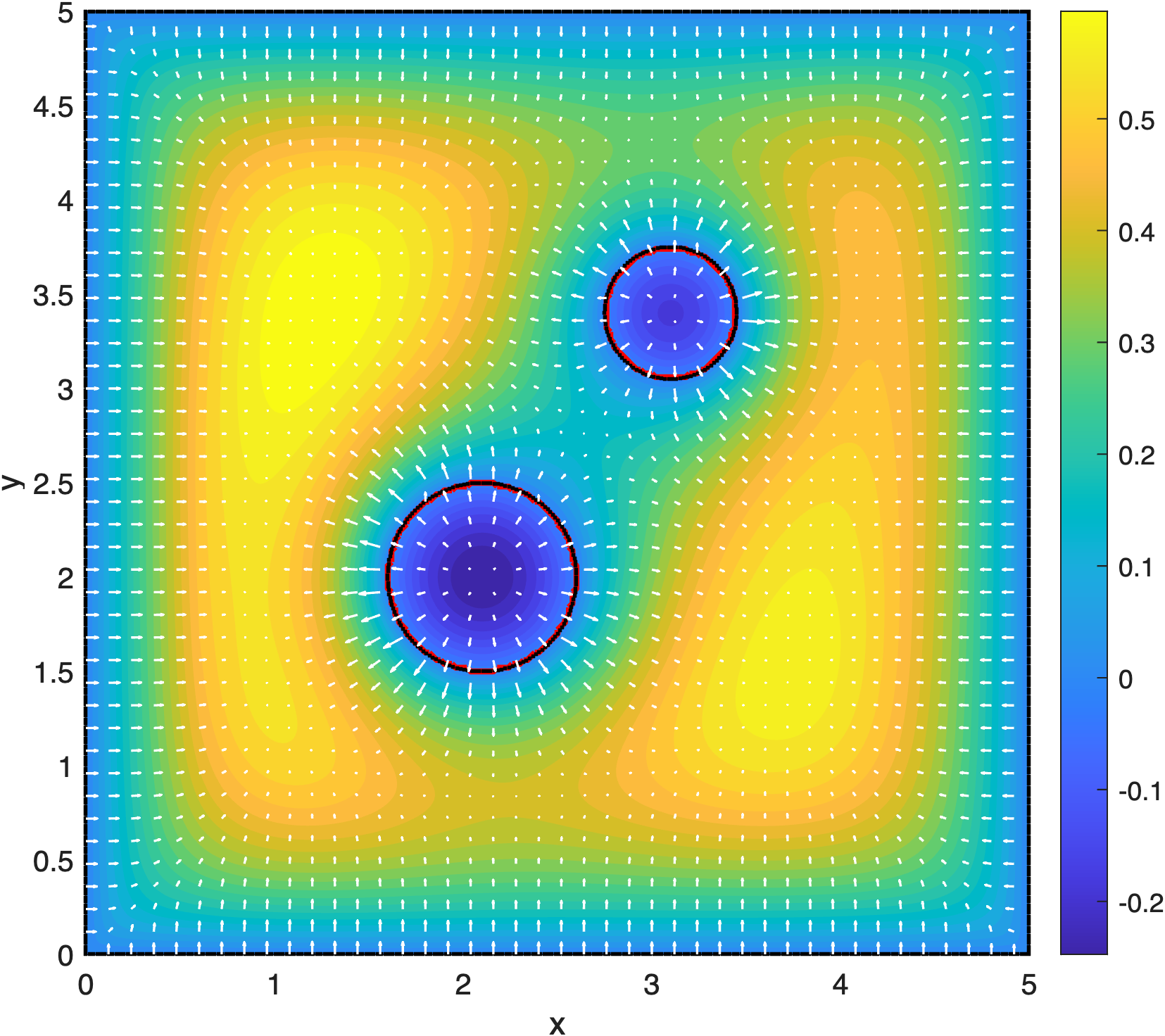}
&
\includegraphics[width=0.31\linewidth]
{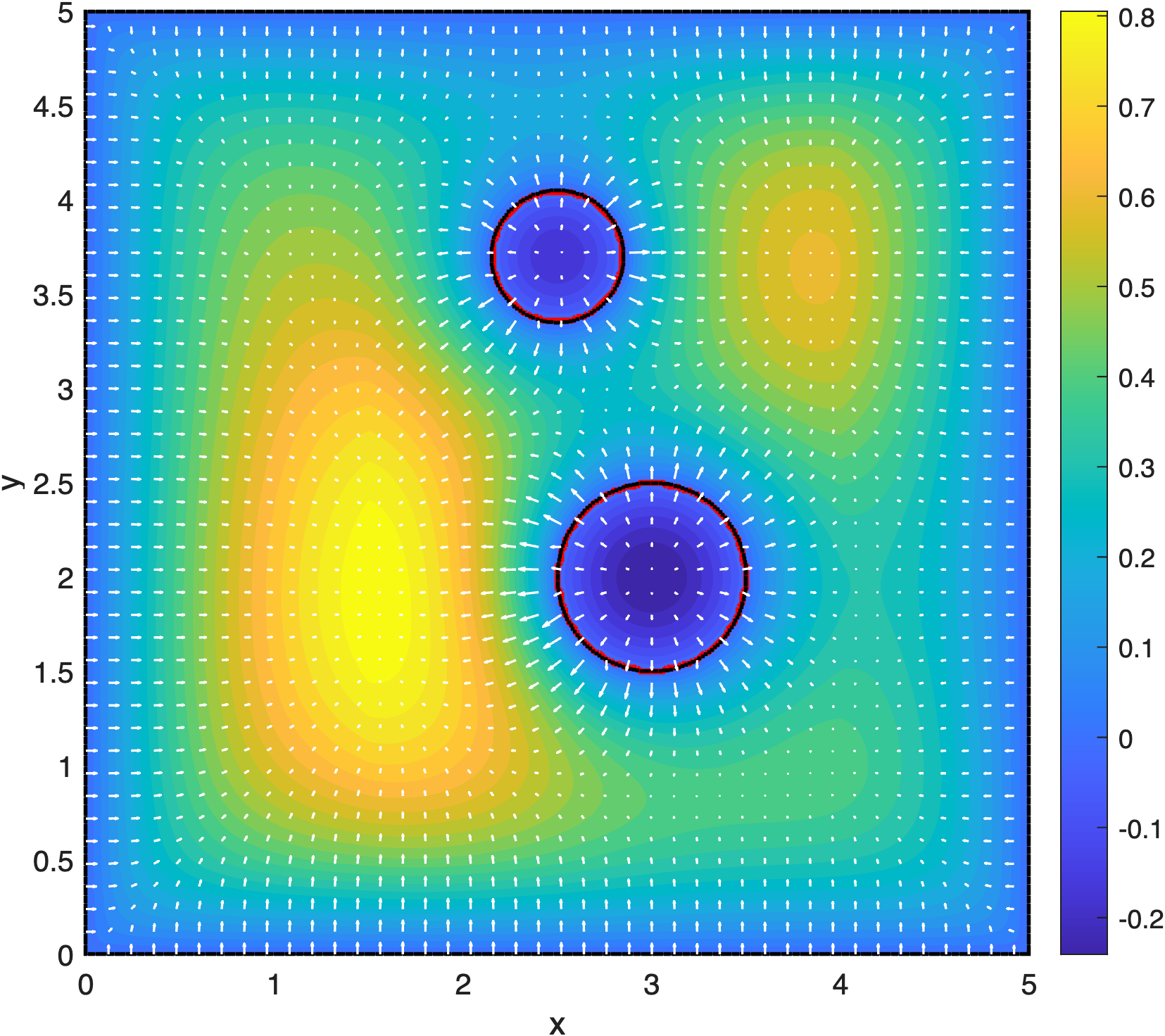}
\\[-1pt]

{\scriptsize (j)} &
{\scriptsize (k)} &
{\scriptsize (l)}

\end{tabular}

\end{tcolorbox}

\end{minipage}

\end{tcolorbox}

\vspace{4pt}

\begin{tcolorbox}[
    width=\textwidth,
    colframe=blue,
    colback=white,
    boxrule=1.1pt,
    arc=1pt,
    left=2pt,
    right=2pt,
    top=2pt,
    bottom=2pt,
    boxsep=0pt
]

\begin{minipage}[t]{0.49\linewidth}
\centering

\begin{tcolorbox}[
    enhanced,
    width=\linewidth,
    colback=white,
    colframe=white,
    boxrule=0pt,
    arc=1pt,
    left=1pt,
    right=1pt,
    top=1pt,
    bottom=1pt,
    boxsep=0pt,
    overlay={
        \draw[
            blue!55,
            line width=0.6pt,
            dash pattern=on 3pt off 2pt
        ]
        (frame.south west) rectangle (frame.north east);
    }
]

\centering
\setlength{\tabcolsep}{1pt}

\begin{tabular}{@{}ccc@{}}

\includegraphics[width=0.31\linewidth]
{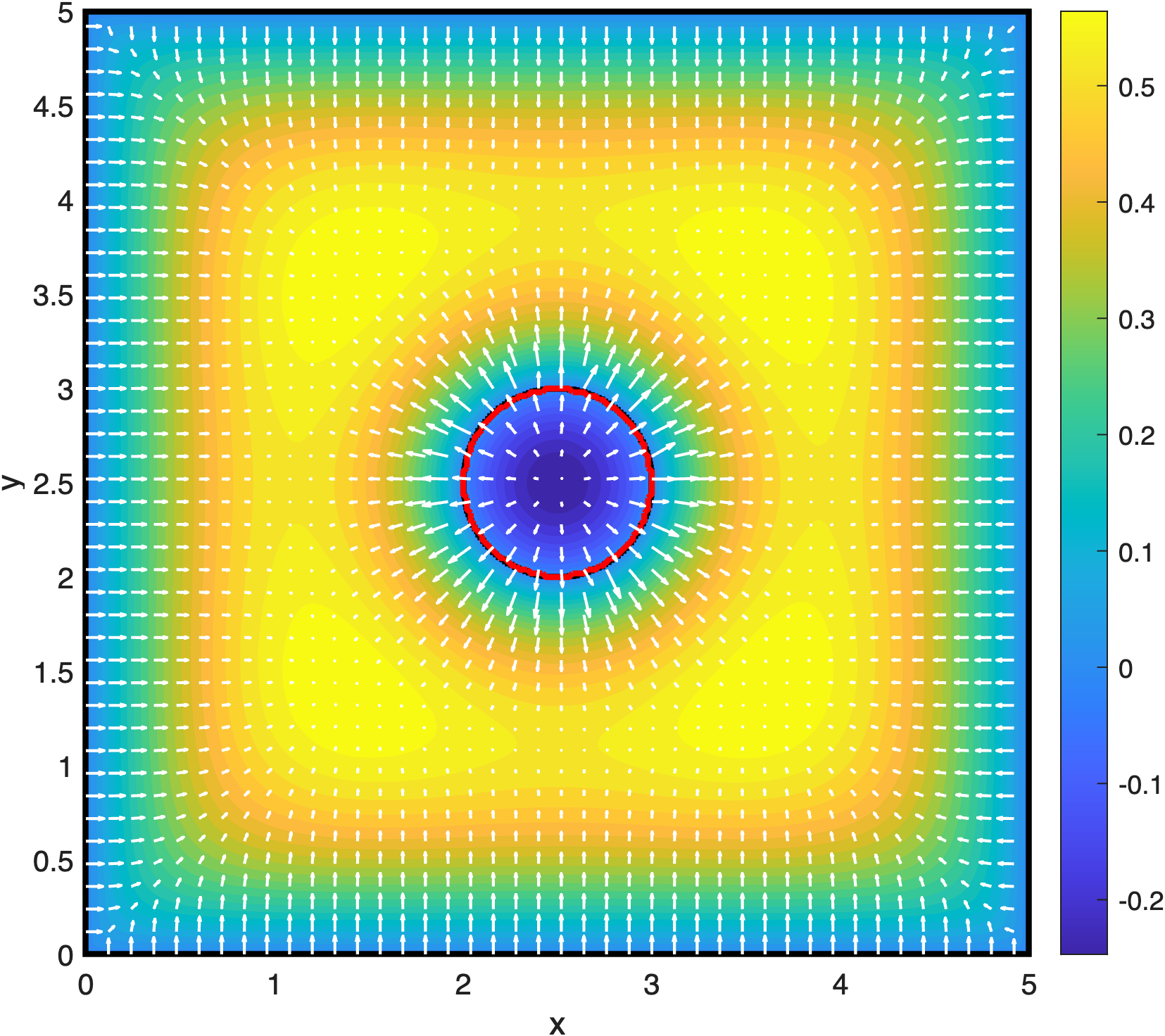}
&
\includegraphics[width=0.31\linewidth]
{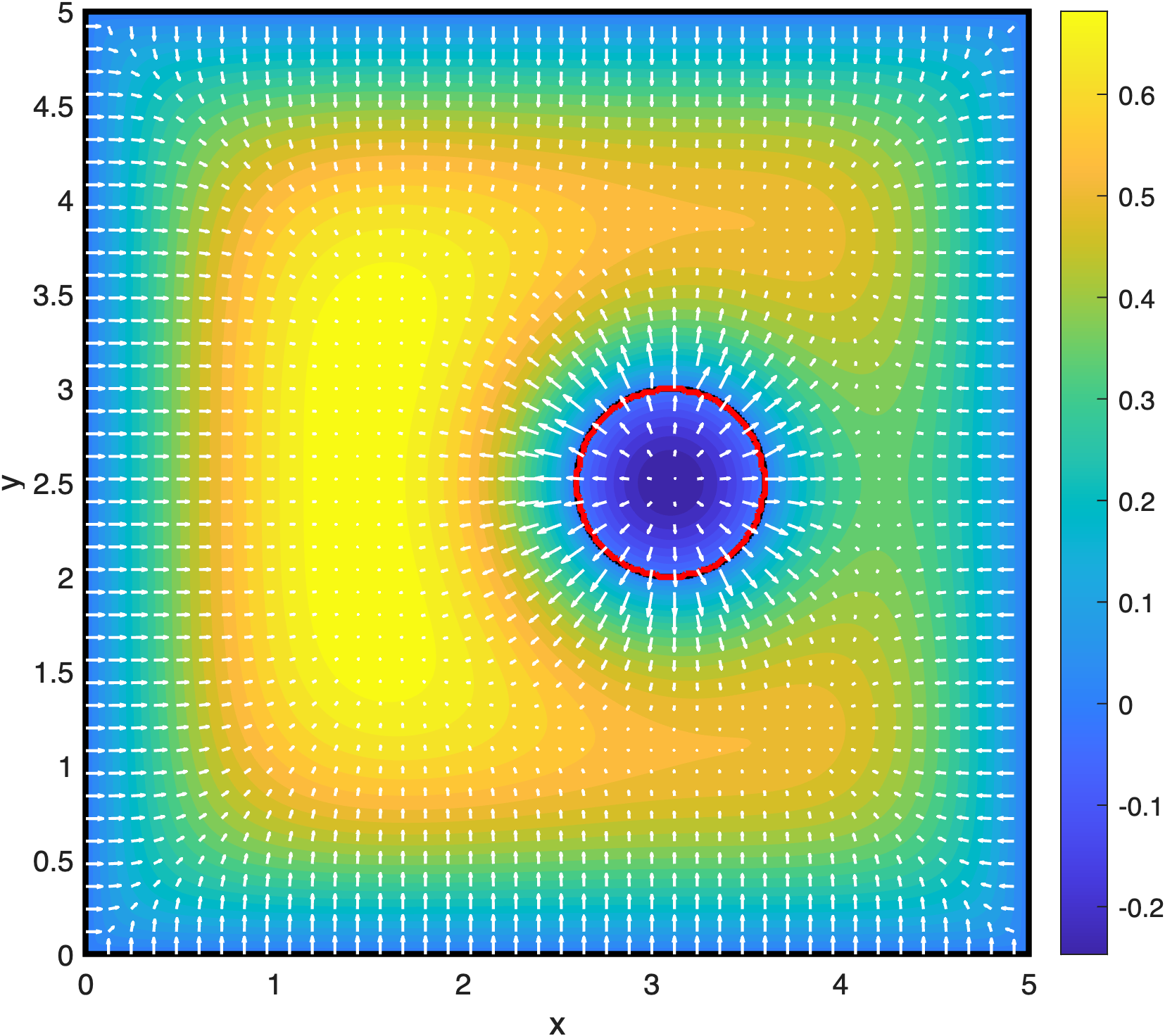}
&
\includegraphics[width=0.31\linewidth]
{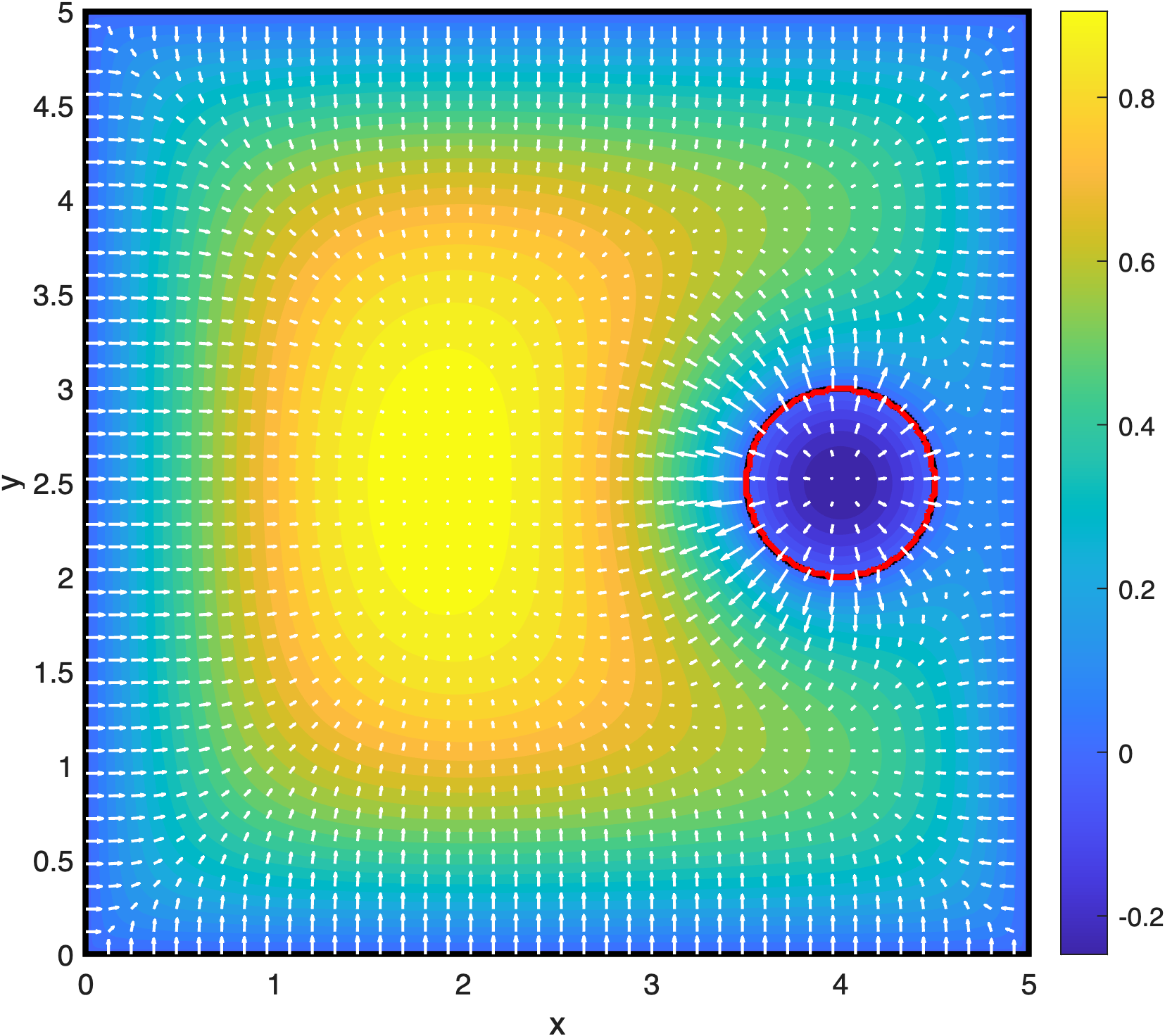}
\\[-1pt]

{\scriptsize (m)} &
{\scriptsize (n)} &
{\scriptsize (o)}

\end{tabular}

\end{tcolorbox}

\end{minipage}%
\hfill%
\begin{minipage}[t]{0.49\linewidth}
\centering


\begin{tcolorbox}[
    enhanced,
    width=\linewidth,
    colback=white,
    colframe=white,
    boxrule=0pt,
    arc=1pt,
    left=1pt,
    right=1pt,
    top=1pt,
    bottom=1pt,
    boxsep=0pt,
    overlay={
        \draw[
            blue!55,
            line width=0.6pt,
            dash pattern=on 3pt off 2pt
        ]
        (frame.south west) rectangle (frame.north east);
    }
]

\centering
\setlength{\tabcolsep}{1pt}

\begin{tabular}{@{}ccc@{}}

\includegraphics[width=0.31\linewidth]
{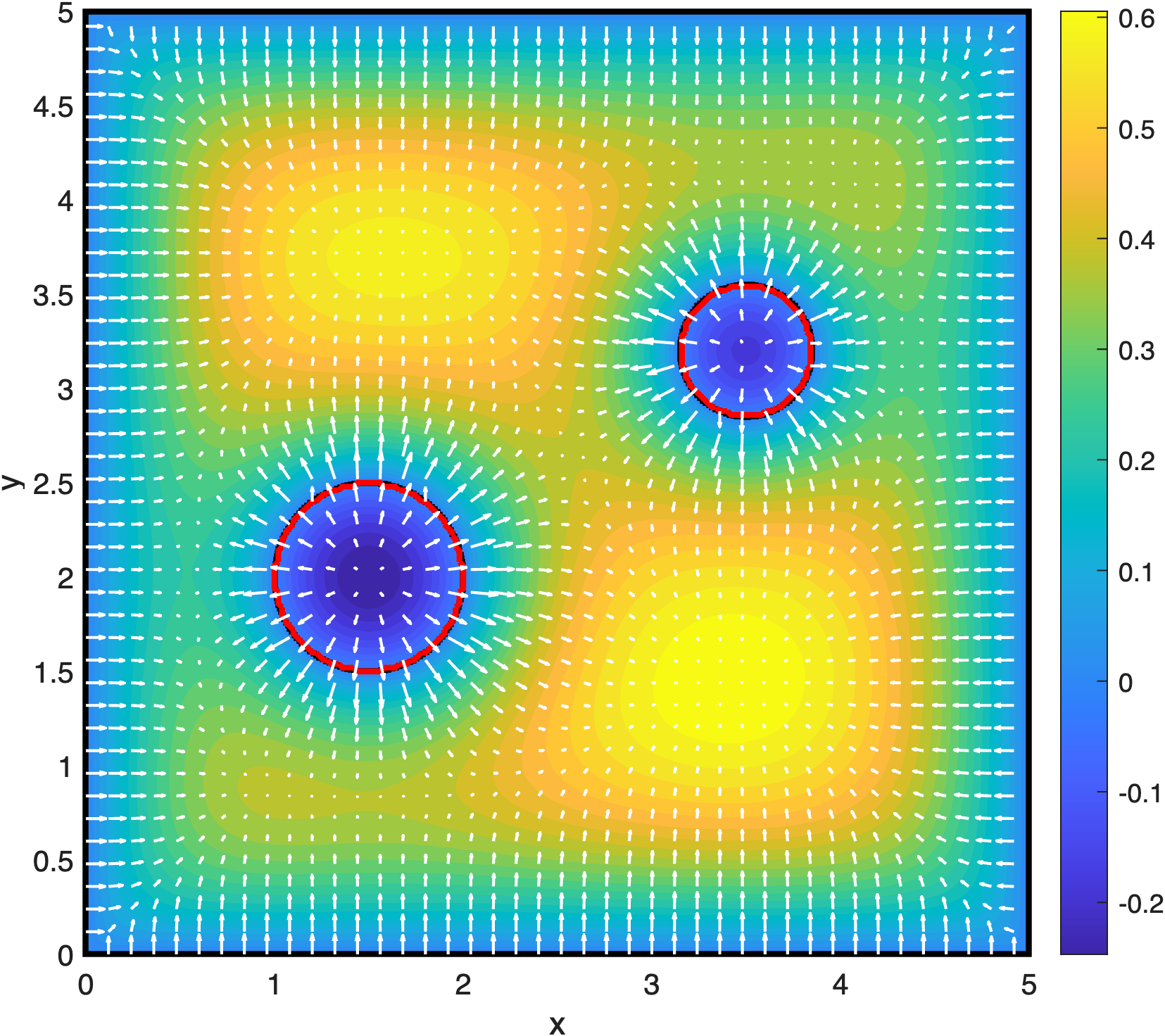}
&
\includegraphics[width=0.31\linewidth]
{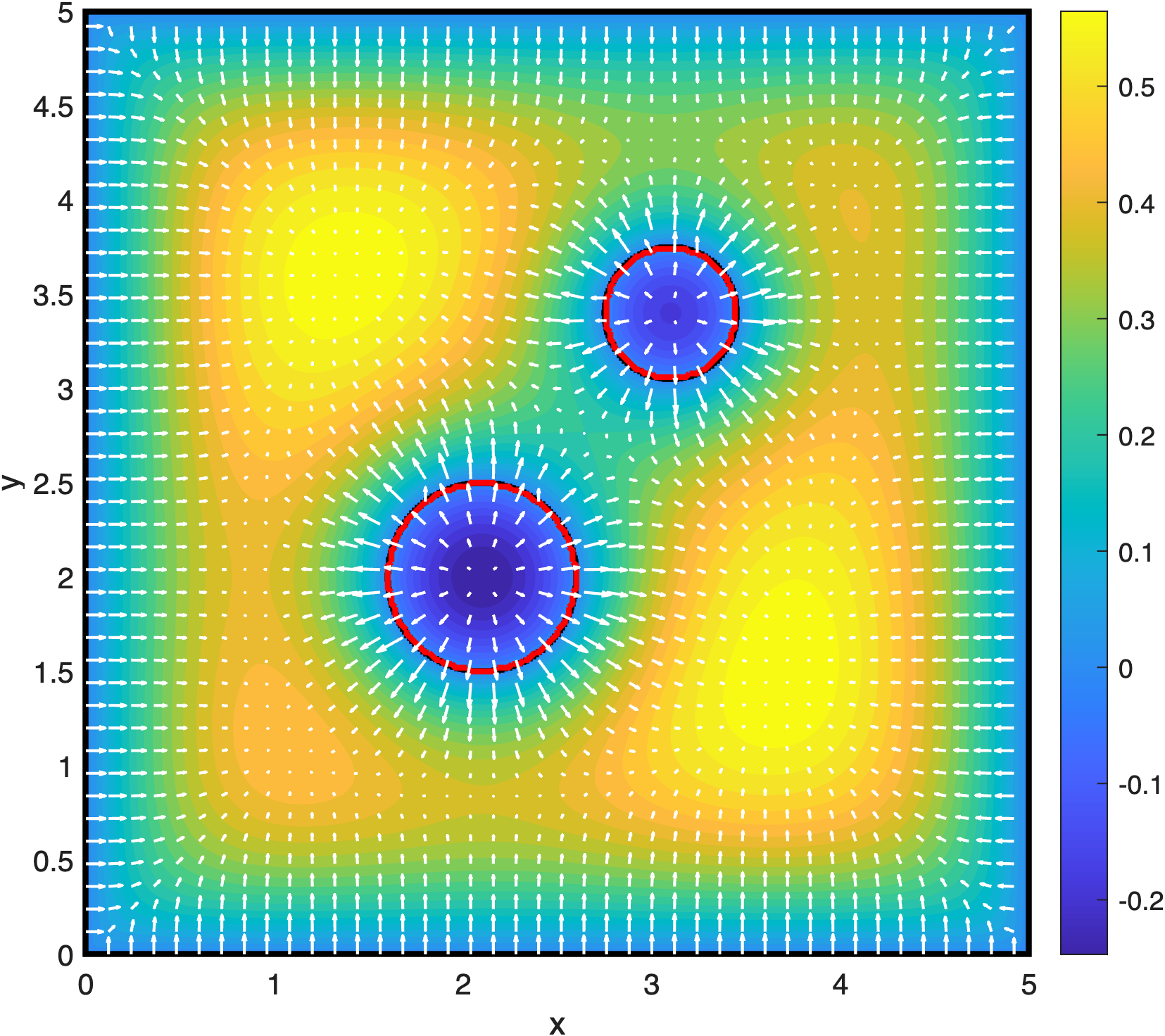}
&
\includegraphics[width=0.31\linewidth]
{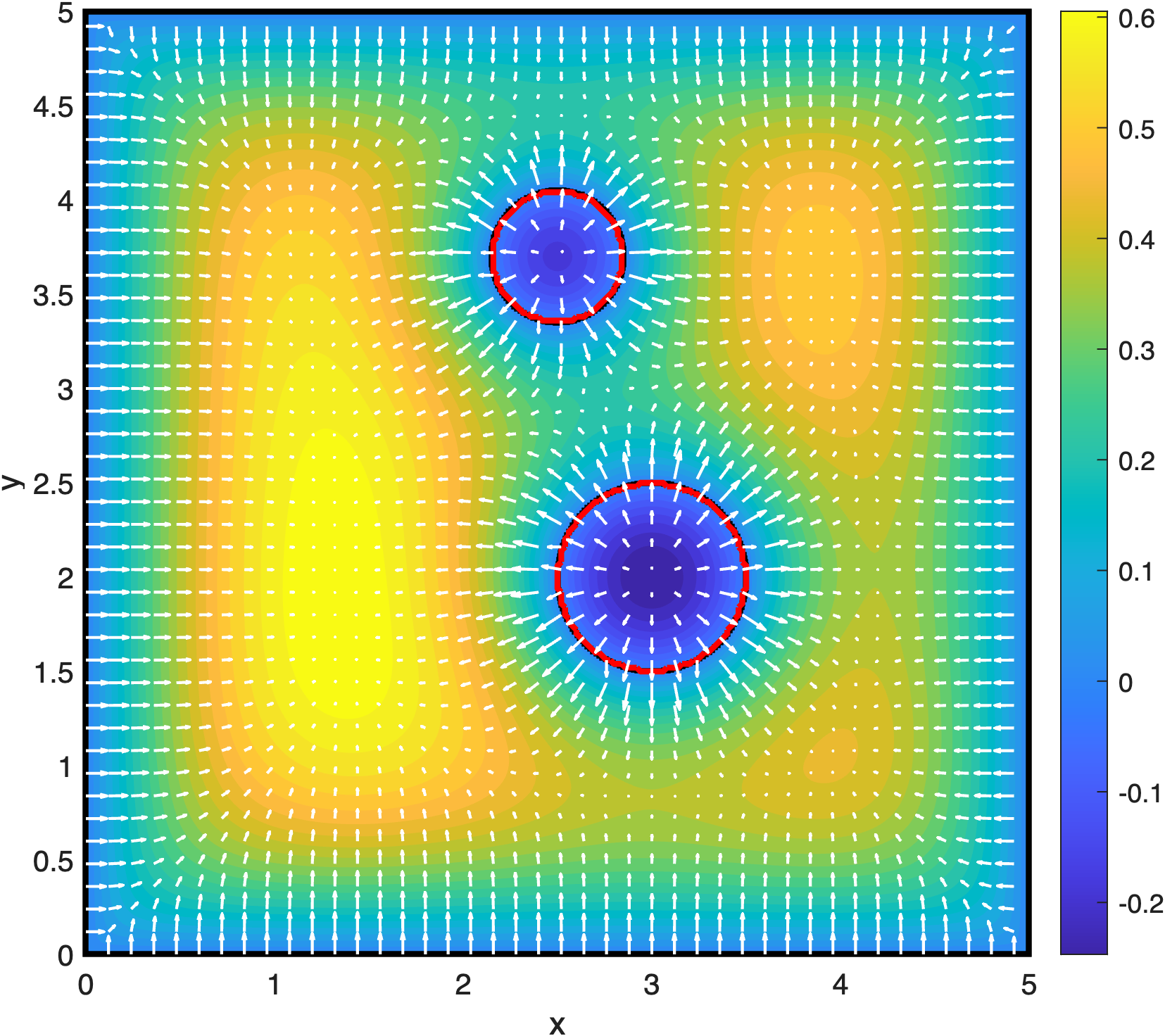}
\\[-1pt]

{\scriptsize (p)} &
{\scriptsize (q)} &
{\scriptsize (r)}

\end{tabular}

\end{tcolorbox}

\end{minipage}

\end{tcolorbox}

\caption{
Comparison of time-varying safety-field evolution over a
$T=5\,\mathrm{s}$  with $dt=0.1\,\mathrm{s}$ horizon for single- and two-obstacle scenarios
with constant obstacle velocities.
The \textbf{red box} shows the proposed transport-driven safety-field update.
Panels (a)--(c) and (g)--(i) correspond to the direct transport step
for the single- and two-obstacle cases, respectively, while
panels (d)--(f) and (j)--(l) show the corresponding safety fields
after the correction step.
The \textbf{blue box} shows the full PDE Re-solve method,
with panels (m)--(o) and (p)--(r) corresponding to the single-
and two-obstacle cases, respectively.
}
\label{fig:fig1}

\end{figure*}

Each residual field is then transported according to its corresponding obstacle motion field $\mathbf{F}_i(\mathbf{x},t)$, following the transport equation in Definition~\ref{def:transport_definition},
\begin{align}\label{eq:trans}
\frac{\partial h_{\mathrm{res}}^{(i)}(\mathbf{x},t)}{\partial t}
+
\mathbf{F}_i(\mathbf{x},t)\cdot
\nabla h_{\mathrm{res}}^{(i)}(\mathbf{x},t)
=
0,
\end{align}
where, $i=1,\ldots,N_{\mathrm{obs}}.$
In this work, the obstacle motion is assumed to be rigid, and the residual field is transported consistently with the corresponding rigid-body motion.
The resulting field $h_{\mathrm{pred}}(\mathbf{x},t) = h_{\mathrm{dom}}(\mathbf{x}) + h_{\mathrm{res}}(\mathbf{x},t)$ does not, in general, satisfy the zero boundary condition on $\partial\Omega(t)$, as illustrated in the first row of Fig.~\ref{fig:fig1}. Therefore, a harmonic correction field $h_{\mathrm{cor}}(\mathbf{x},t)$ is introduced
\begin{align}\label{eq:correction}
\begin{cases}
\Delta h_{\mathrm{cor}}(\mathbf{x},t)=0,
& \mathbf{x}\in\Omega(t),\\
h_{\mathrm{cor}}(\mathbf{x},t)
=
- h_{\mathrm{pred}}(\mathbf{x},t),
& \mathbf{x}\in\partial\Omega(t).
\end{cases}
\end{align}
By construction, the resulting field $h(\mathbf{x},t) = h_{\mathrm{pred}}(\mathbf{x},t) + h_{\mathrm{cor}}(\mathbf{x},t)$ satisfies $h(\mathbf{x},t)=0$ on $\partial\Omega(t)$.
As illustrated in the second row of Fig.~\ref{fig:fig1},  the required boundary condition is therefore restored after the correction is applied.
When compared with the full re-solving procedure in Algorithm~\ref{alg:baseline_resolve}, the proposed method in Algorithm~\ref{alg:transport_correction} avoids repeatedly reconstructing the boundary-normal field, guidance vector field, and forcing function. Therefore, at each time step, only a transport update and a harmonic correction are required.

\begin{remark}
Rather than reconstructing the full boundary-normal field, guidance
vector field, and forcing function at every time step as required by the
Re-solve baseline \cite{bena2025geometry, Gilbert2025dynamic}, the
proposed method advances $h_{\mathrm{res}}^{(i)}$ purely by transport and
applies a single harmonic correction to restore the boundary condition.
As shown in Fig.~\ref{fig:fig1}, for obstacles moving at constant, possibly different, velocities over $T=5\,\mathrm{s}$ with $dt=0.1\,\mathrm{s}$, this yields corrected fields at $t=5\,\mathrm{s}$ (panels (f), (l)) that are visually indistinguishable from the Re-solve results (panels (o), (r)) in both safety value and gradient structure, while requiring substantially less computation, as quantified in Sections~\ref{sec:simulation} and~\ref{sec:experiment}.
\end{remark}

\begin{algorithm}[t]
\caption{Proposed Transport-Correction Update}
\label{alg:transport_correction}
\begin{algorithmic}[1]
\Require Initial safety field $h(\mathbf{x},t_0)$, obstacle motion field $\mathbf{F}(\mathbf{x},t)$
\Ensure Time-varying safety field $h(\mathbf{x},t_k)$

\For{$k=1,2,\ldots$}
    \State Update obstacle boundary using~\eqref{eq:transport}
    \State Obtain the predicted field $h_{\mathrm{pred}}(\mathbf{x},t_k)$ by transport equation \ref{eq:trans}
    \State Obtain the correction field $h_{\mathrm{cor}}(\mathbf{x},t_k)$ by solving~\eqref{eq:correction}
    \State Update the safety field
    $h(\mathbf{x},t_k)$
\EndFor
\end{algorithmic}
\end{algorithm}

\section{Predictive Safety Control via MPC-DCBF}\label{sec:control}
The time-varying safety field $h(\mathbf{x},t)$ constructed in the previous section provides a continuous-time safety representation of the dynamic environment. For the control implementation, we evaluate this field at discrete sampling instants and formulate the corresponding time-varying DCBF (TV-DCBF) condition, which is subsequently incorporated into an MPC framework.

\subsection{Time-Varying DCBF Formulation}
At the sampling instants $t_k = k\Delta t$, we denote $h_k(\mathbf{x})= h(\mathbf{x},t_k)$, and define the corresponding time-varying safe set as
$\mathcal{C}_k
=
\left\{
\mathbf{x}\in\mathbb{R}^n
\mid
h_k(\mathbf{x})\geq0
\right\}$.
Consider the discrete-time system
$\mathbf{x}_{k+1} = \mathbf{F}_d(\mathbf{x}_k,\mathbf{u}_k)$.
Building on the standard DCBF condition in Definition~\ref{def2:dcbf}, we extend the barrier condition to the time-varying safety field as
\begin{align}
h_{k+1}\!\left(
\mathbf{F}_d(\mathbf{x}_k,\mathbf{u}_k)
\right)
\geq
\rho\,h_k(\mathbf{x}_k),
\label{eq:tv_dcbf}
\end{align}
where $0 \le \rho \leq 1$. We refer to~\eqref{eq:tv_dcbf} as the TV-DCBF condition. Unlike the standard DCBF condition in Definition~\ref{def2:dcbf}, condition~\eqref{eq:tv_dcbf} relates two generally distinct functions $h_{k+1}$ and $h_k$ at consecutive sampling instants, since the safety field evolves with the obstacle geometry between sampling instants. Consequently, the forward invariance result in Lemma~\ref{lem:dcbf_forward_invariance}, which applies to a fixed barrier function $h$, does not directly apply to the sequence of time-varying safe sets $\mathcal{C}_k$. The following proposition extends this safety guarantee to the time-varying setting.

\begin{proposition}[Safety guarantee under the TV-DCBF Condition]
\label{prop:tv_dcbf_safety}
Suppose that $\mathbf{x}_0\in\mathcal{C}_0$ and that the TV-DCBF condition~\eqref{eq:tv_dcbf} holds for all $k\geq0$. Then
\begin{align}
h_k(\mathbf{x}_k)
\geq
\rho^k h_0(\mathbf{x}_0)
\geq0,
\qquad
\forall k\geq0,
\end{align}
and hence the state trajectory remains within the time-varying safe set $\mathcal{C}_k$ at every sampling instant.
\end{proposition}
\begin{proof}
From~\eqref{eq:tv_dcbf}, we have
\begin{align}
h_{k+1}(\mathbf{x}_{k+1})
\geq
\rho\,h_k(\mathbf{x}_k).
\end{align}
Applying this inequality recursively yields
\begin{align}
h_k(\mathbf{x}_k)
\geq
\rho^k h_0(\mathbf{x}_0).
\end{align}
Since $\mathbf{x}_0\in\mathcal{C}_0$, we have
$h_0(\mathbf{x}_0)\geq0$. Together with $0<\rho\leq1$, this gives
$h_k(\mathbf{x}_k)\geq0$ for all $k\geq0$, and hence
$\mathbf{x}_k\in\mathcal{C}_k$ for all $k\geq0$.
\end{proof}

\subsection{MPC with Time-Varying DCBF Constraints}
At each sampling instant $k$, the TV-DCBF condition~\eqref{eq:tv_dcbf} is incorporated into the MPC problem as a safety constraint over the prediction horizon. Let $N$ denote the prediction horizon, and let $\hat{\mathbf{x}}_{i|k}$ and $\hat{\mathbf{u}}_{i|k}$ denote the predicted state and control input $i$ steps ahead from sampling instant $k$. The MPC problem is formulated as
\begin{align}\label{eq:mpc-cbf}
\min_{\{\hat{\mathbf{u}}_{i|k}\}_{i=0}^{N-1}}
\quad &
\sum_{i=0}^{N-1}
\ell(\hat{\mathbf{x}}_{i|k},\hat{\mathbf{u}}_{i|k})
+
\ell_f(\hat{\mathbf{x}}_{N|k})
\\
\text{s.t.}\quad
&
\hat{\mathbf{x}}_{0|k}=\mathbf{x}_k,\nonumber
\\
&
\hat{\mathbf{x}}_{i+1|k}
=
\mathbf{F}_d(
\hat{\mathbf{x}}_{i|k},
\hat{\mathbf{u}}_{i|k}),\nonumber
\\
&
h_{k+i+1}(\hat{\mathbf{x}}_{i+1|k})
\geq
\rho\,h_{k+i}(\hat{\mathbf{x}}_{i|k}),
\label{eq:mpc_tvdcbf}
\\
&
\hat{\mathbf{u}}_{i|k}\in\mathcal{U}.\nonumber
\end{align}
where $i=0,\ldots,N-1$.
Unlike a static MPC-CBF formulation, the safety constraint \eqref{eq:mpc_tvdcbf} is evaluated using the time-indexed safety fields $h_{k+i}$ over the prediction horizon. These fields are obtained by applying the transport-and-correction procedure in Section~\ref{sec:time-varying cbf} to the predicted obstacle boundary at each future sampling instant, assuming a known obstacle motion model. This allows the predicted robot trajectory to be evaluated against the safety field corresponding to each future sampling instant, without requiring the full PDE to be re-solved at every predicted step.

At each sampling instant, only the first control input
$\hat{\mathbf{u}}_{0|k}^{\star}$ of the optimal sequence is applied to the system, and the optimisation is repeated at the next sampling instant according to the receding-horizon principle. Provided that the MPC problem remains feasible and the applied control input satisfies the TV-DCBF constraint at each sampling instant, Proposition~\ref{prop:tv_dcbf_safety} guarantees that the closed-loop state remains within the corresponding time-varying safe set.

\section{Evaluation of Dynamic Safety Fields in Simulation}\label{sec:simulation}
This section evaluates the computational efficiency of the proposed safety field update method, comparing it against the Re-solve method in both two-dimensional (2D) and three-dimensional (3D) environments. Since the focus here is on field update performance, the controller is not included in the simulations.

\subsection{Safety Field Update}\label{sec:simulation-safety}
\begin{figure*}[htbp]
    \centering
    \rule{0pt}{12pt}\par

    \begin{subfigure}[t]{0.30\textwidth}
        \centering
        \includegraphics[width=\linewidth]
        {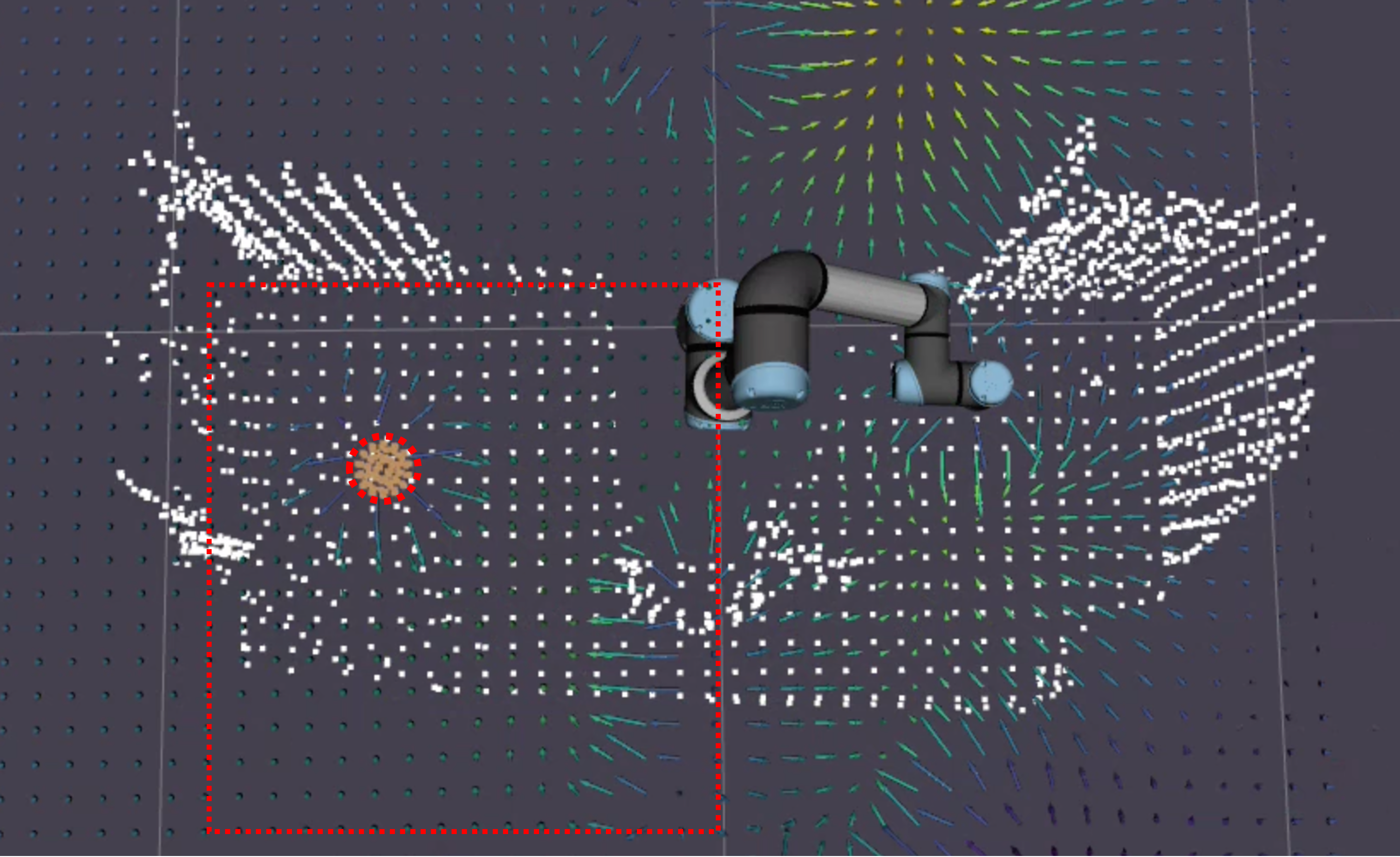}
        \caption{}
    \end{subfigure}
    \begin{subfigure}[t]{0.17\textwidth}
        \centering
        \includegraphics[width=\linewidth]
        {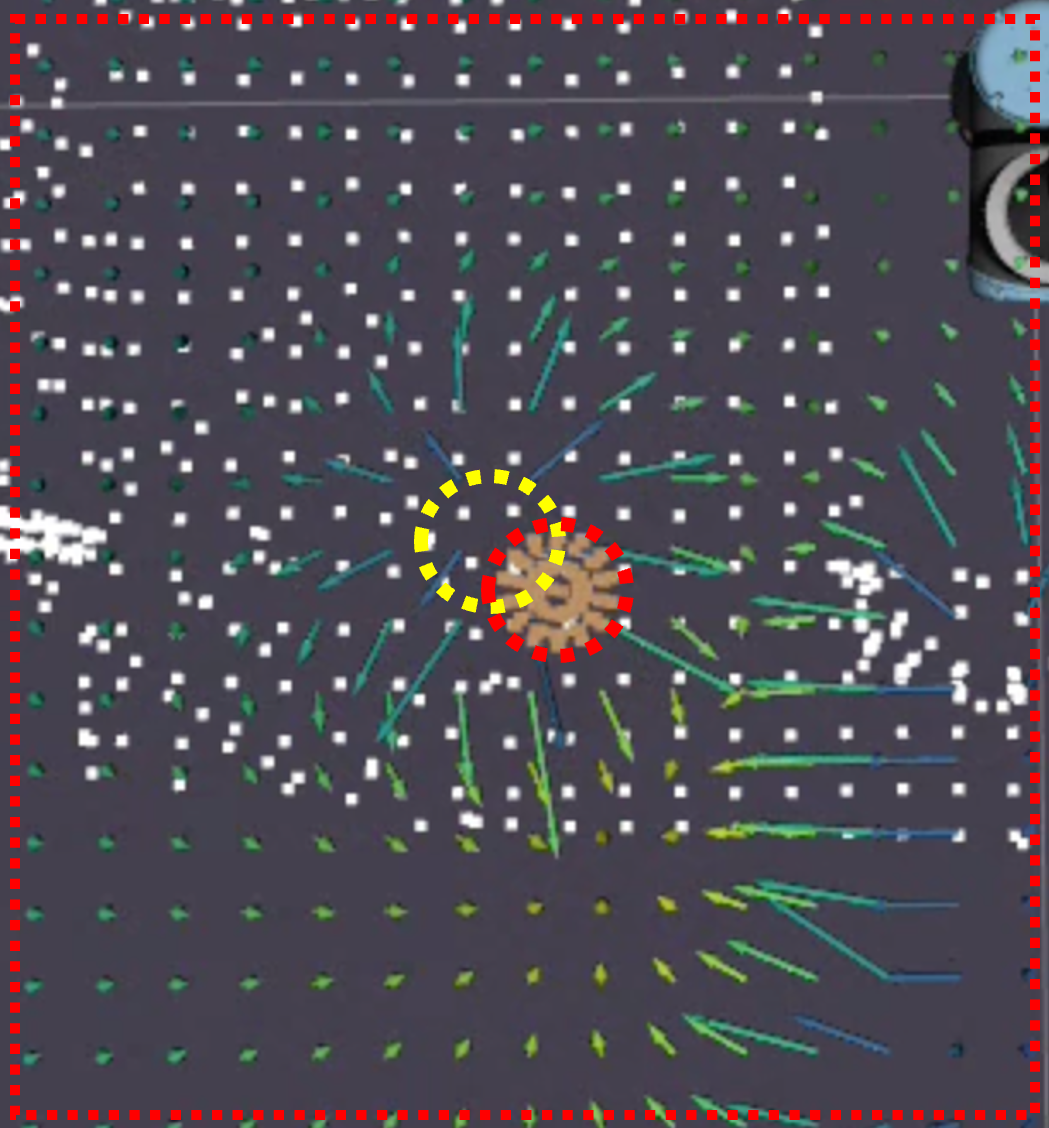}
        \caption{}
    \end{subfigure}
    \begin{subfigure}[t]{0.17\textwidth}
        \centering
        \includegraphics[width=\linewidth]
        {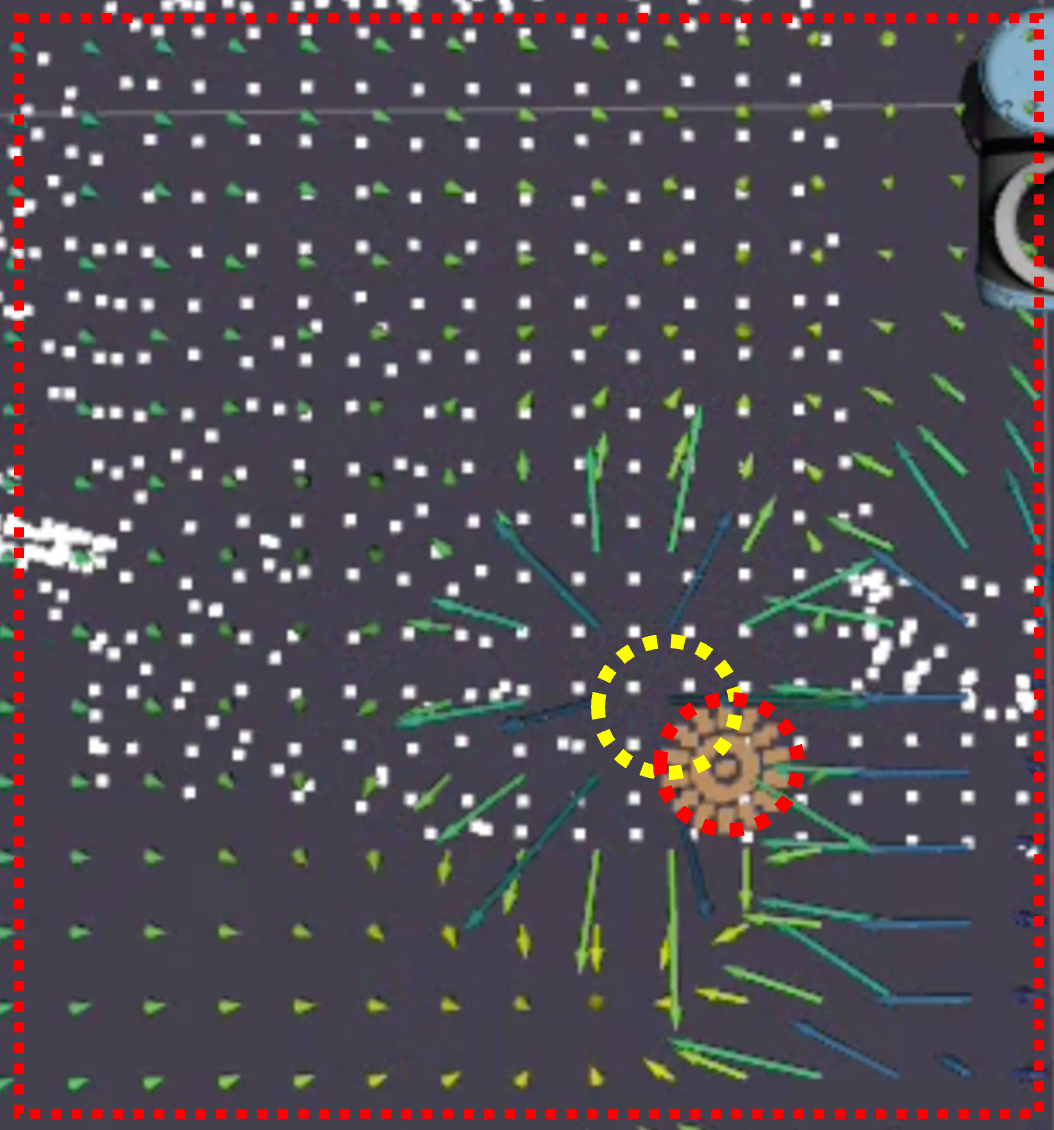}
        \caption{}
    \end{subfigure}
    \begin{subfigure}[t]{0.30\textwidth}
        \centering
        \includegraphics[width=\linewidth]
        {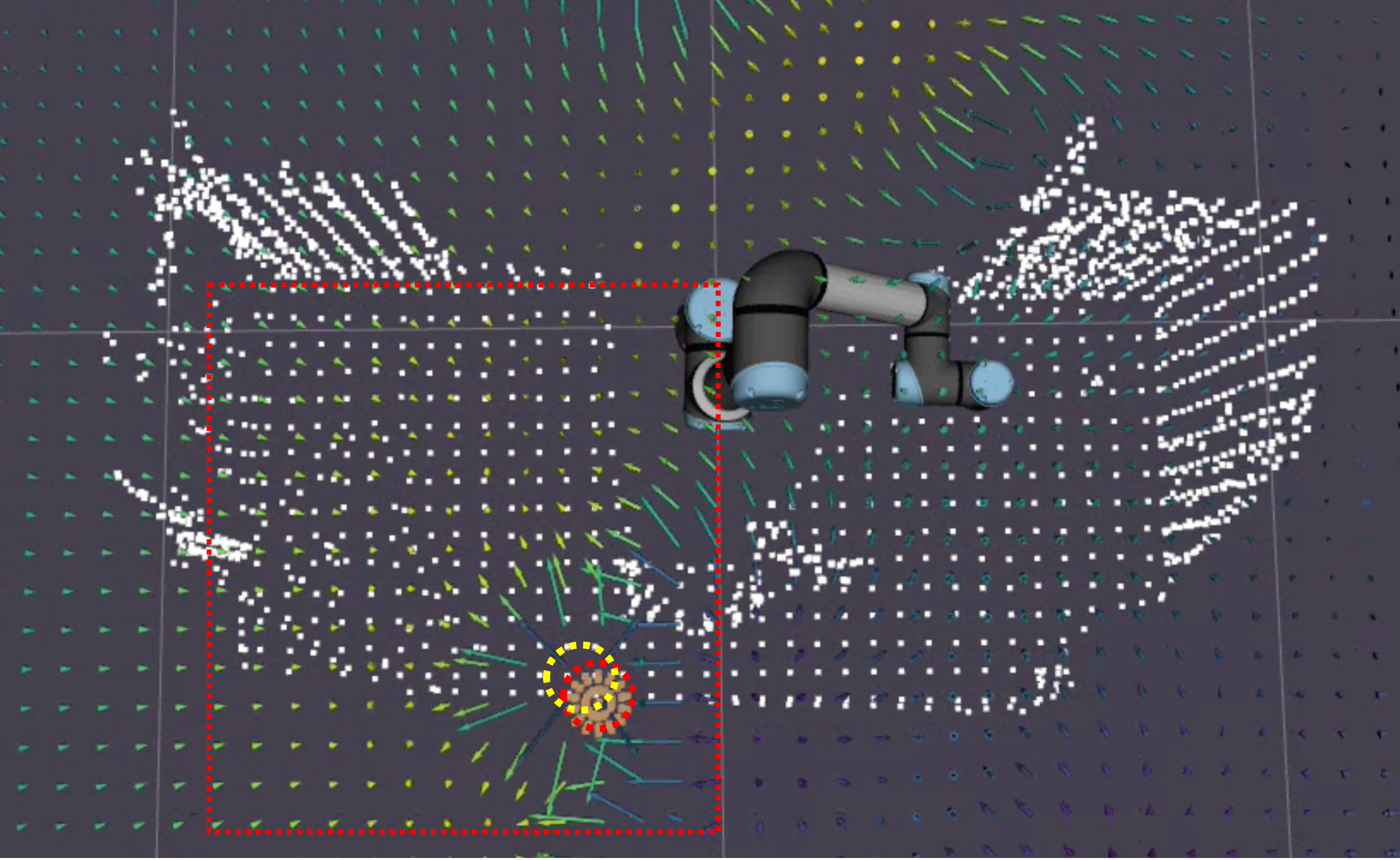}
        \caption{}
    \end{subfigure}

    \vspace{2mm}

    \begin{subfigure}[t]{0.30\textwidth}
        \centering
        \includegraphics[width=\linewidth]
        {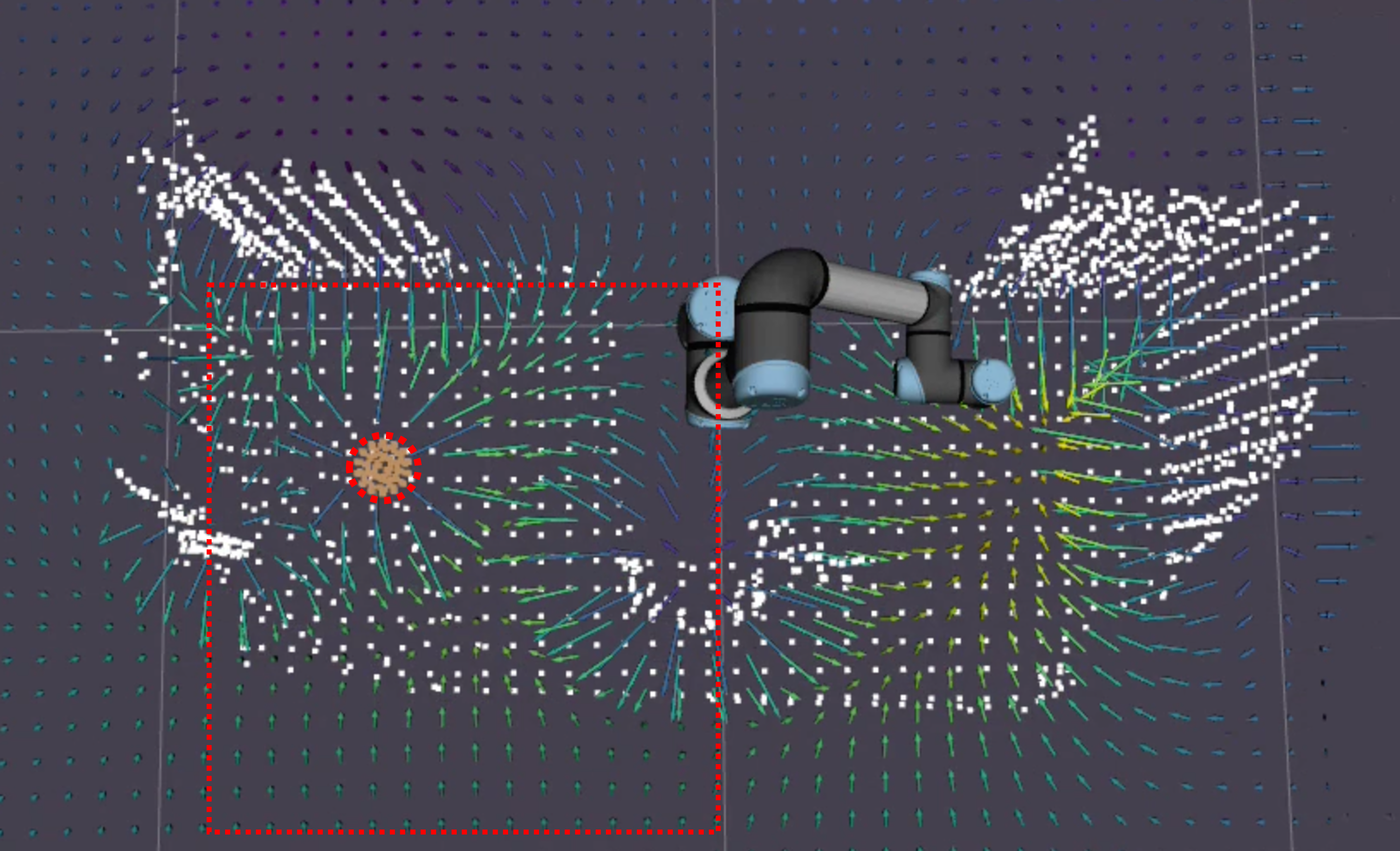}
        \caption{}
    \end{subfigure}
    \begin{subfigure}[t]{0.17\textwidth}
        \centering
        \includegraphics[width=\linewidth]
        {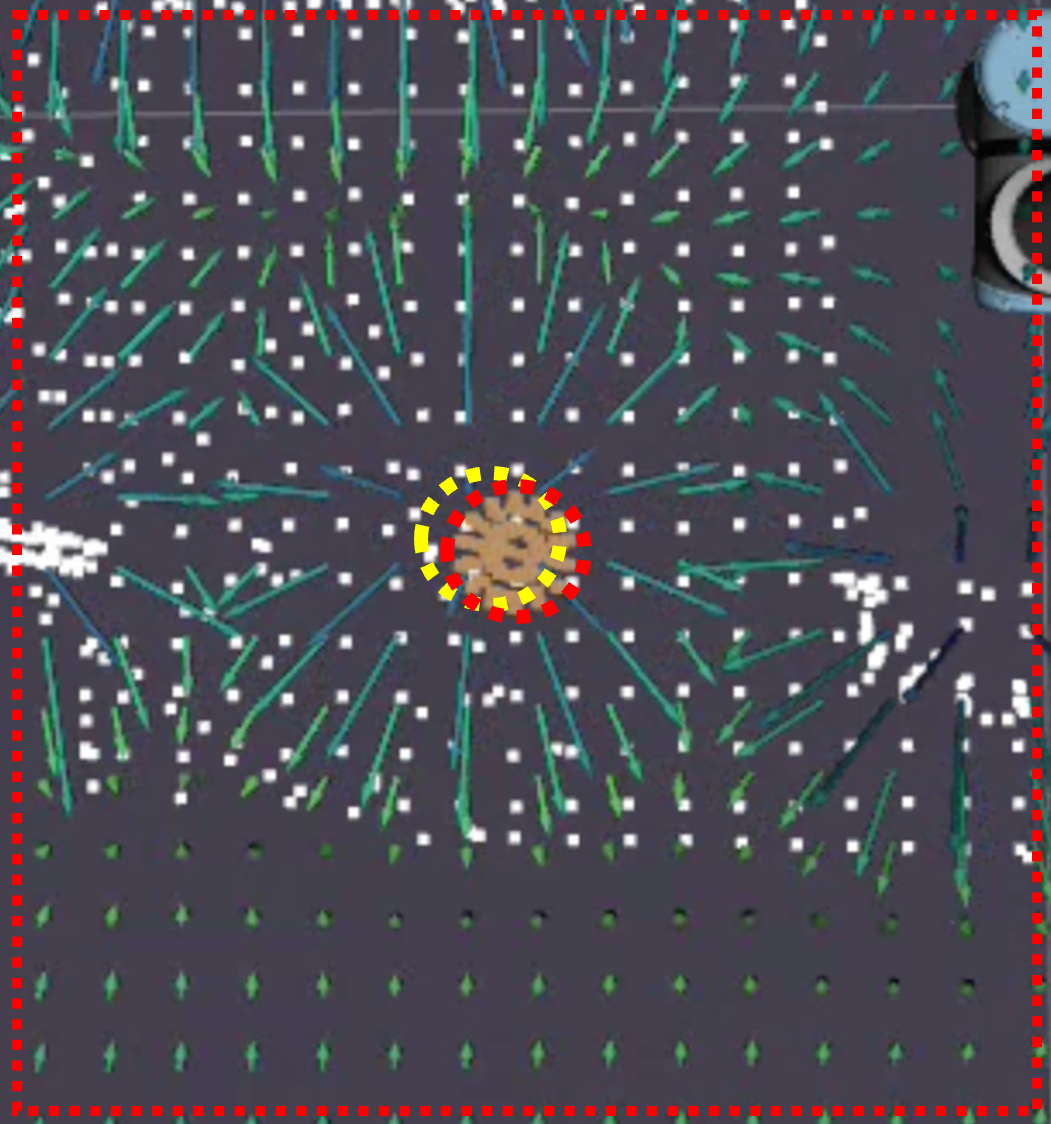}
        \caption{}
    \end{subfigure}
    \begin{subfigure}[t]{0.17\textwidth}
        \centering
        \includegraphics[width=\linewidth]
        {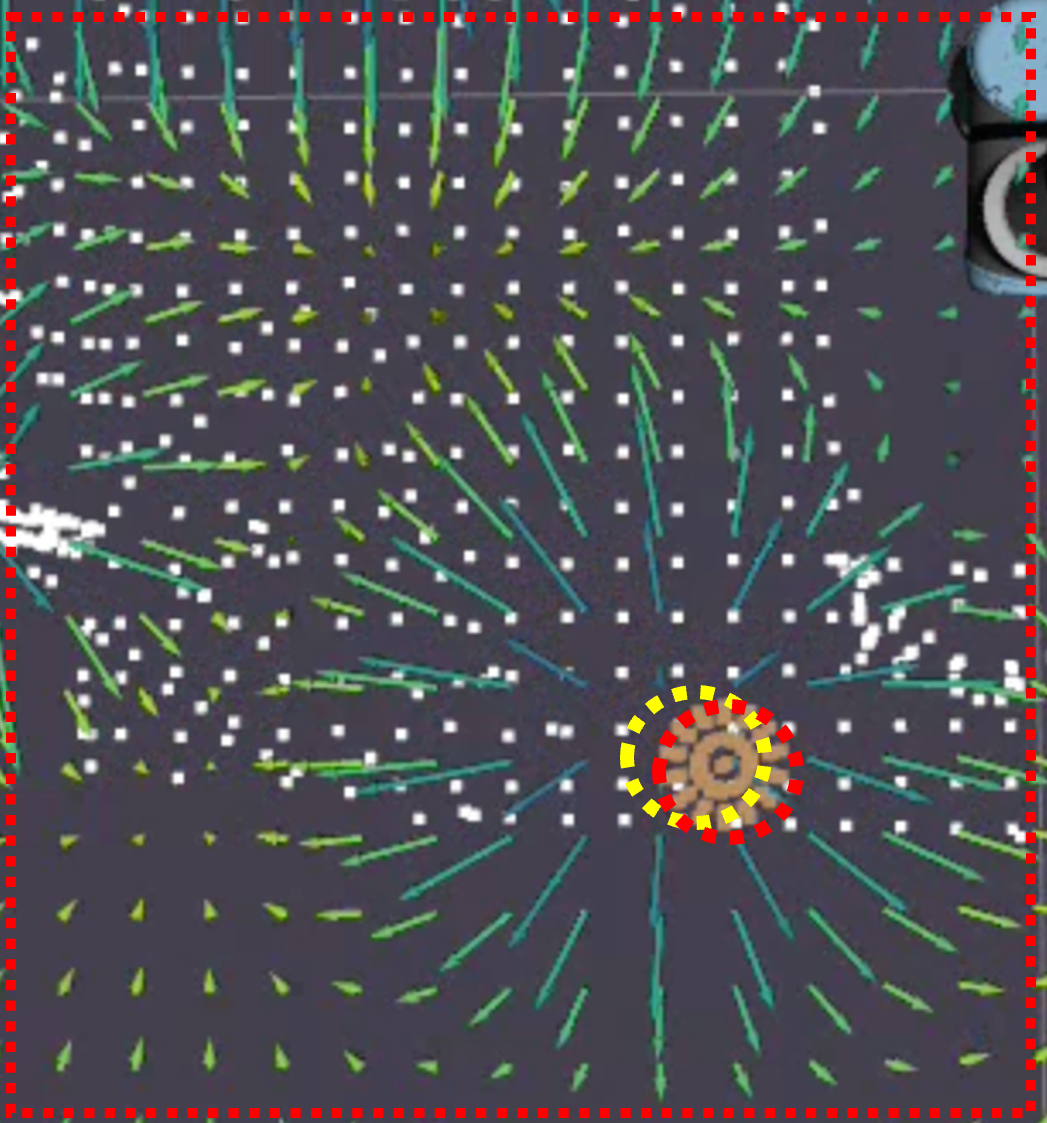}
        \caption{}
    \end{subfigure}
    \begin{subfigure}[t]{0.30\textwidth}
        \centering
        \includegraphics[width=\linewidth]
        {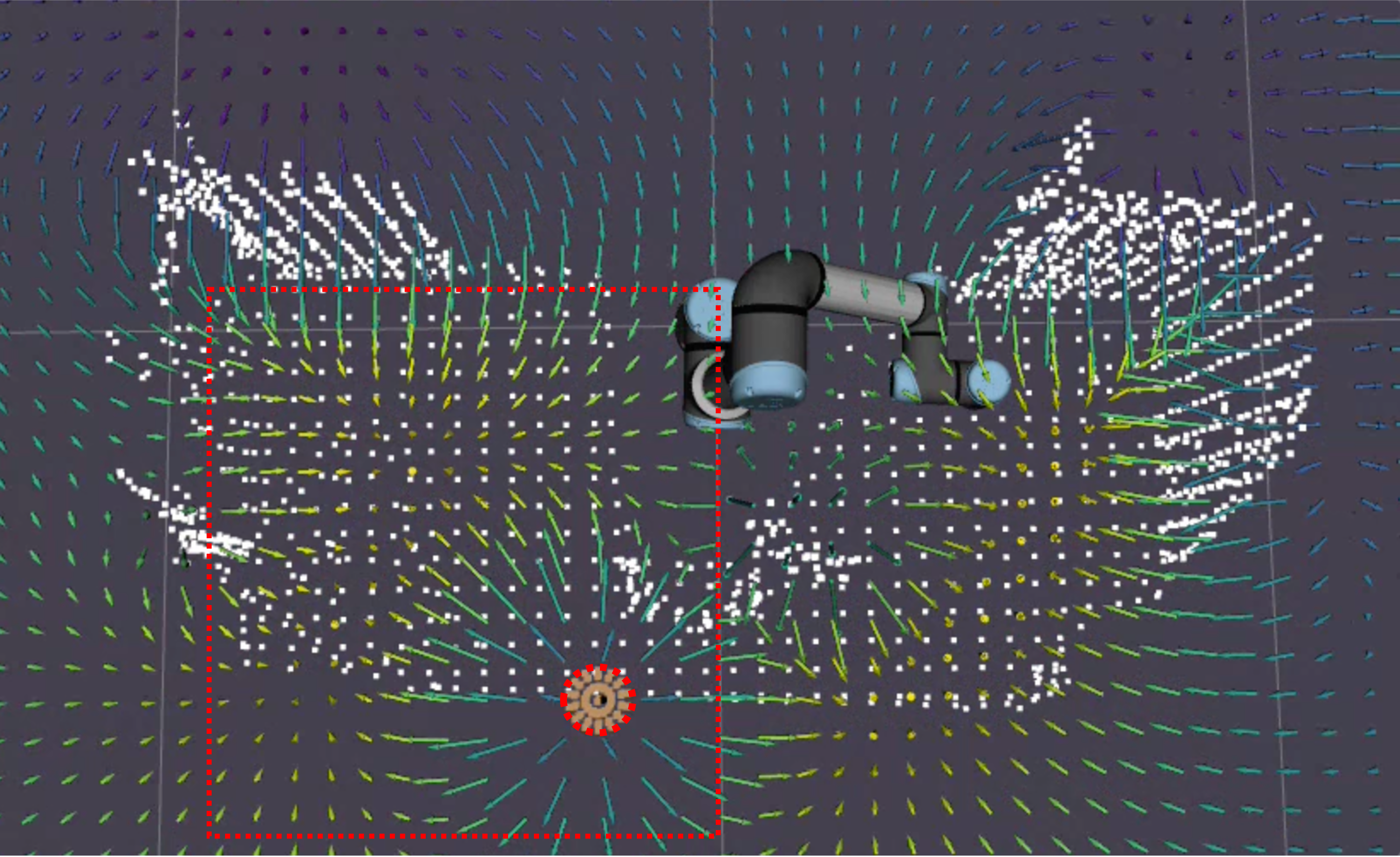}
        \caption{}
    \end{subfigure}

    \caption{Safety-field evolution: Re-solve method in~\cite{bena2025geometry, Gilbert2025dynamic} (top) vs. proposed method (bottom) at four time instants. The red-boxed region, identical in (a) and (e), is magnified in (b) and (c) for the Re-solve method and in (f) and (g) for the Transport-driven method, enabling a clearer comparison between the zero-level set of the safety field (yellow dashed line) and the current obstacle position (red dashed line).}
    \label{fig:h_update_comparison}
\end{figure*}

To evaluate the update efficiency in a 3D environment, we construct a real-world workspace and use a camera to capture its geometry as a point cloud, from which the safety field is generated. 
The computational domain is discretised using a $45\times45\times45$ grid over $x\in[-1.0,1.5]$ m, $y\in[-1.5,1.0]$ m, and $z\in[0,1.5]$ m. The PDEs are solved using successive over-relaxation (SOR) with a relaxation factor of $\omega=1.9$, a convergence tolerance of $10^{-5}$, and a maximum of 3000 iterations, which are identical for both methods.
Within this workspace, we consider a spherical moving obstacle with a radius of $0.045$ m with an initial centre at $[0.25,-0.60,0.20]$ m and a constant velocity of $[0.20,0.20,0.20]$ m/s. 
The obstacle state is updated at 20 Hz over a 10 s interval, and both the proposed method and the Re-solve method are evaluated under the same workspace and obstacle motion.

\begin{table}[htbp]
\centering
\caption{Update efficiency comparison in the 3D experiment.}
\label{tab:update_time_3d}
\begin{tabular}{lccc}
\hline
Method & Avg. Updates & Avg. time (s) & Max. time (s) \\
\hline
Re-solve & 12 & 1.6347 & 2.4252 \\
Ours     & \textbf{27} & \textbf{0.5805} & \textbf{1.0029} \\
\hline
\end{tabular}
\end{table}

To provide an intuitive comparison, we first present a representative case illustrating how the safety field evolves as the obstacle moves. 
Fig.~\ref{fig:h_update_comparison} compares the Re-solve method in panels (a)–(d) with the proposed method in panels (e)–(h), covering the initial obstacle position in (a) and (e), two intermediate positions in (b), (c), (f) and (g), and the final position (d) and (h). To more clearly show the safety-field evolution during the intermediate stages, panels (b), (c), (f), and (g) are enlarged within the red-boxed regions.
As summarised in Table~\ref{tab:update_time_3d}, the proposed method performs 27 safety-field updates over the 10 s interval, with an average update time of 0.58 s. This higher update frequency allows the safety field to more closely follow the obstacle's motion, as illustrated in panels (f) and (g). In contrast, the Re-solve method performs only 12 updates over the same interval, with a much larger average update time of 1.63 s, resulting in a noticeable delay between the obstacle motion and the corresponding safety-field evolution, as observed in panels (b) and (c).

\subsection{Randomised Evaluation}

\begin{figure}[htbp]
    \centering

    \begin{subfigure}[t]{0.125\textwidth}
        \centering
        \includegraphics[width=\linewidth]{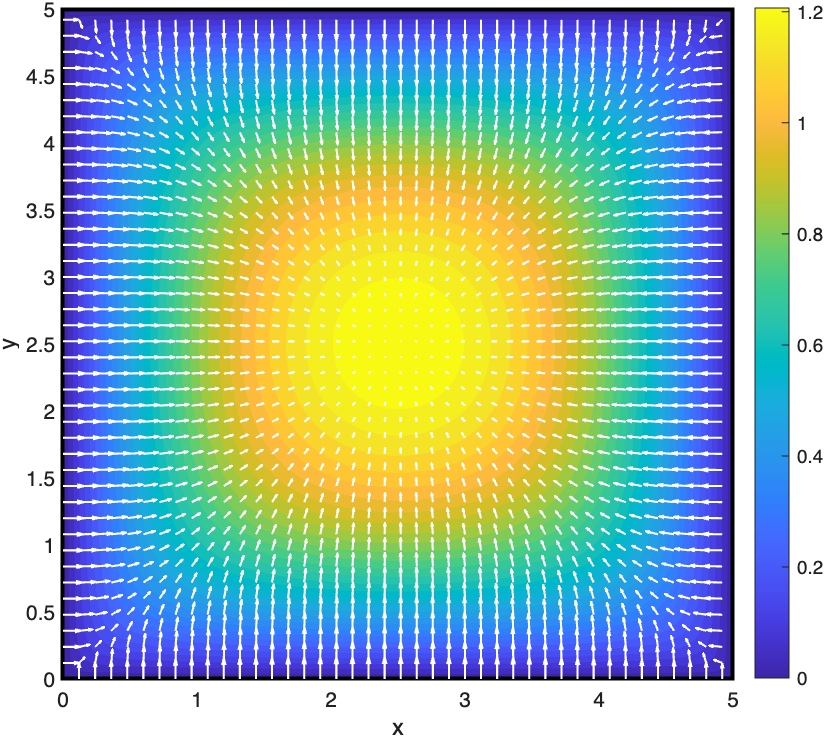}
        \caption{}
    \end{subfigure}
    \hfill
    \begin{subfigure}[t]{0.13\textwidth}
        \centering
        \includegraphics[width=\linewidth]{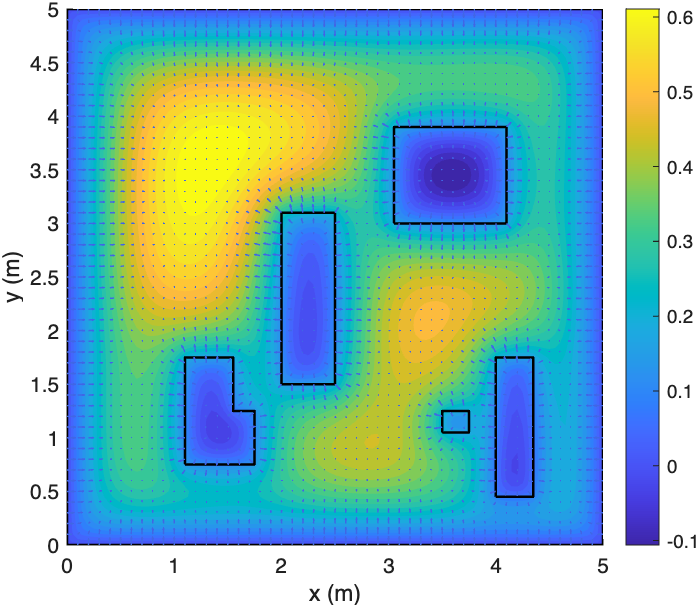}
        \caption{}
    \end{subfigure}
    \hfill
    \begin{subfigure}[t]{0.15\textwidth}
        \centering
        \includegraphics[width=\linewidth]{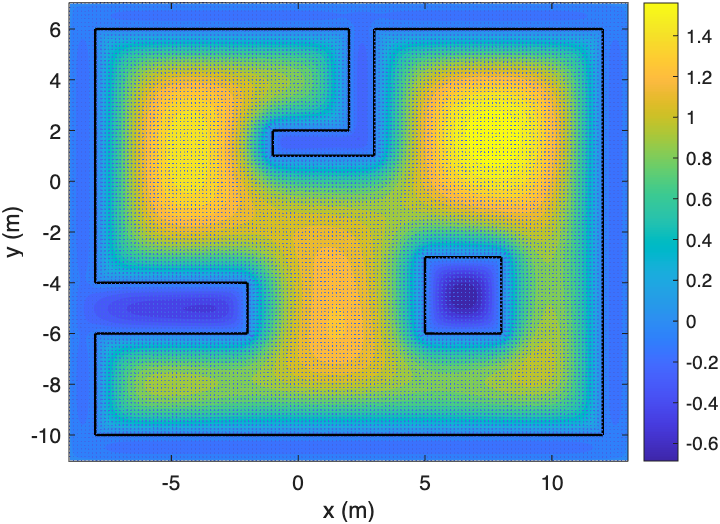}
        \caption{}
    \end{subfigure}

    \vspace{0.5em}

    \begin{subfigure}[t]{0.15\textwidth}
        \centering
        \includegraphics[width=\linewidth]{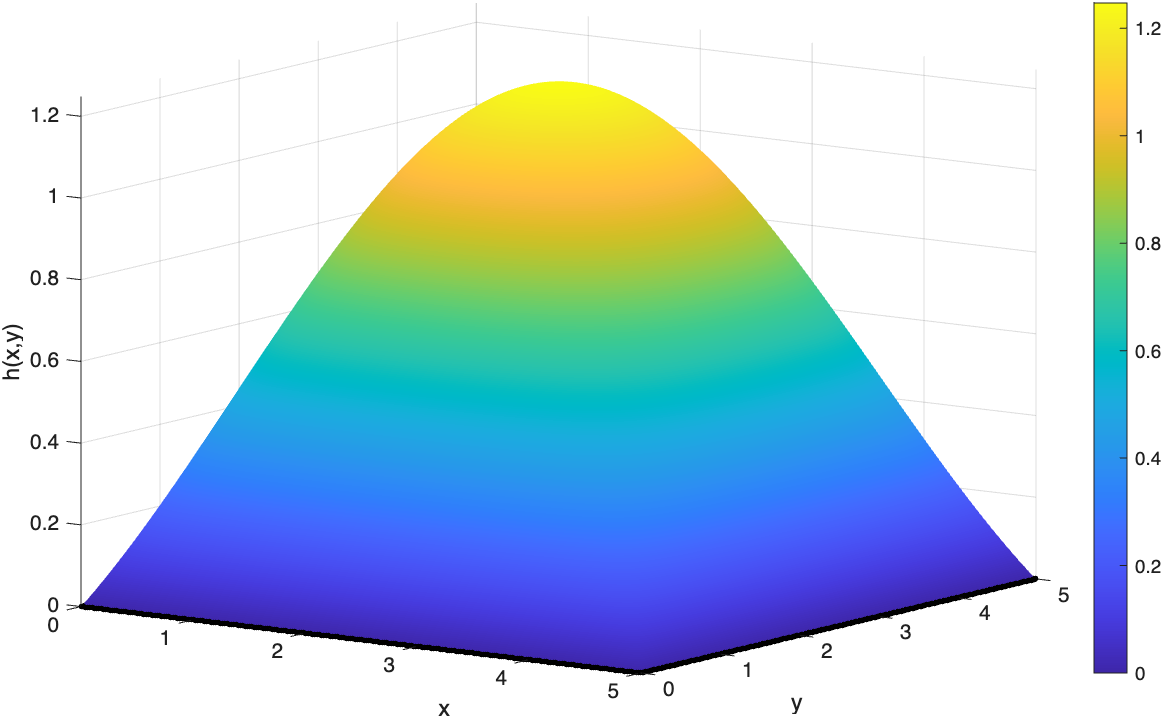}
        \caption{}
    \end{subfigure}
    \hfill
    \begin{subfigure}[t]{0.15\textwidth}
        \centering
        \includegraphics[width=\linewidth]{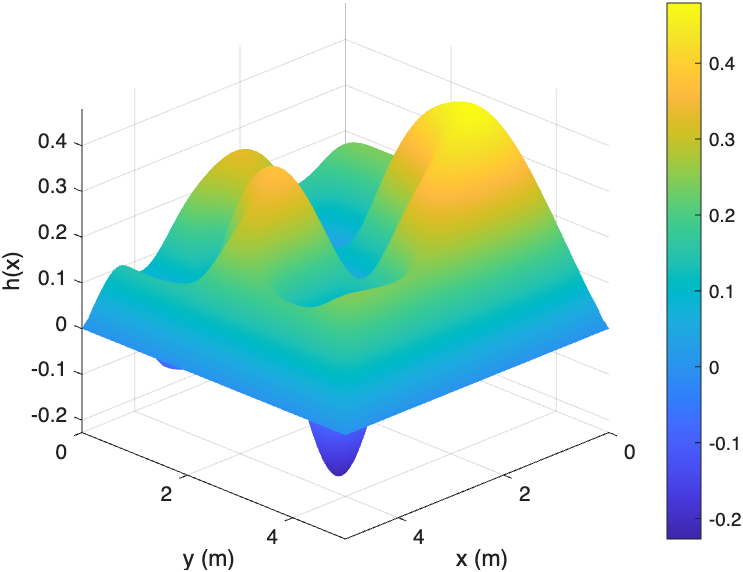}
        \caption{}
    \end{subfigure}
    \hfill
    \begin{subfigure}[t]{0.15\textwidth}
        \centering
        \includegraphics[width=\linewidth]{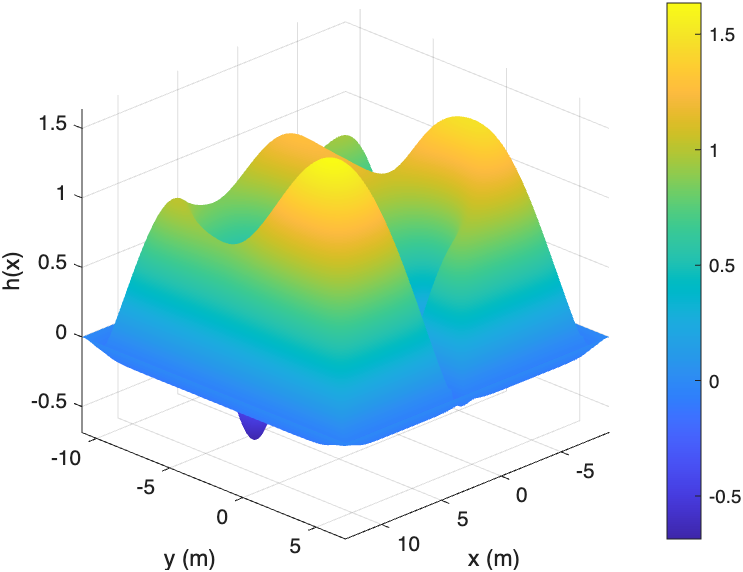}
        \caption{}
    \end{subfigure}

    \caption{2D simulation environments for Maps 1--3: 2D safety fields and gradients (top row) and corresponding 3D safety fields (bottom row).}
    \label{fig:maps}
\end{figure}

To provide a more comprehensive quantitative evaluation, we further compare the two methods using a randomly generated moving obstacle in both 2D and 3D environments. 
For the 2D evaluation, three different environments are considered, as shown in Fig.~\ref{fig:maps} and denoted as Map 1, Map 2, and Map 3. For the 3D evaluation, the same real-world workspace described in Section~\ref{sec:simulation-safety} is used.
For each environment, the moving obstacle's initial position and velocity are randomly generated, and the obstacle state is updated at 20 Hz over a horizon of $T=10$ s.
The same SOR solver settings described in Section~\ref{sec:simulation-safety} are used, with a $251\times251$ grid for the 2D environments and a $45\times45\times45$ grid for the 3D environment.
A total of 100 moving-obstacle trials are randomly generated for each environment, and the same set of trials is used for both the proposed method and the Re-solve method. This ensures that any performance difference reflects the update method itself rather than variations in obstacle trajectories.

We report the average number of safety field updates completed within the 10 s horizon (Avg. updates), together with the average, minimum, and maximum computation time required for a single safety field update in Table~\ref{tab:random_update_results}.
Across all environments, the proposed method completes substantially more updates than the Re-solve method in~\cite{bena2025geometry, Gilbert2025dynamic} within the same horizon, with around 12 to 25 times more updates in the 2D maps and around 3 times more in the 3D environment. 
This advantage becomes more pronounced as the free space increases, since the Re-solve method must perform SOR iterations over a larger number of free grid points, whereas the proposed method avoids repeatedly solving the full PDE from scratch.
Correspondingly, the average update time of the proposed method is significantly lower, achieving speedups of around 20 to 45 times in the 2D maps and around 10 times in the 3D environment. 
Notably, even the maximum observed update time of the proposed method remains lower than the minimum observed update time of the Re-solve method across all environments, showing that the proposed method consistently maintains a lower update time throughout the tested cases.
Since the safety field is subsequently used to provide safety information to the controller, timely updates are critical for responding to obstacle motion and maintaining effective safe control. This benefit is further demonstrated through the control experiments in the next section \ref{sec:experiment}.

\begin{table*}[htbp]
\centering
\rule{0pt}{12pt}\par
\caption{Safety-field update performance over 100 random moving-obstacle trials with $T=10$ s.}
\label{tab:random_update_results}
\resizebox{\textwidth}{!}{
\begin{tabular}{cl|cccc|cccc}
\toprule
\multicolumn{2}{c|}{\multirow{2}{*}{Environment}}
& \multicolumn{4}{c|}{Ours}
& \multicolumn{4}{c}{Re-solve in~\cite{bena2025geometry, Gilbert2025dynamic}} \\
\cline{3-6} \cline{7-10}
& & Frequency (Hz)
& Avg. time (s)
& Min. time (s)
& Max. time (s)
& Frequency (Hz)
& Avg. time (s)
& Min. time (s)
& Max. time (s) \\
\midrule
\multirow{3}{*}{2D}
& Map 1
& \textbf{121.36} & \textbf{0.0404} & 0.0193 & 0.2313
& 4.81 & 1.8263 & 1.4332 & 2.1477 \\
& Map 2
& \textbf{159.35} & \textbf{0.0310} & 0.0142 & 0.0822
& 13.34 & 0.6754 & 0.5424 & 0.8158 \\
& Map 3
& \textbf{167.00} & \textbf{0.0283} & 0.0049 & 0.1291
& 13.70 & 0.6545 & 0.6080 & 0.9847 \\
\midrule
3D
& Point cloud
& \textbf{153.88} & \textbf{0.0185} & 0.0064 & 0.0709
& 45.17 & 0.1782 & 0.1713 & 0.2396 \\
\bottomrule
\end{tabular}
}
\end{table*}

\section{Real-World Experiments with Predictive Control}\label{sec:experiment}

\subsection{System and Controller}
In the real-world experiments, we consider safe motion of the end-effector (EE) of a Universal Robots UR5e manipulator. The EE position is defined as
$\mathbf{x}=[x,y,z]^{\top}\in\mathbb{R}^{3}$,
and the control input
$\mathbf{u}=[u_x,u_y,u_z]^{\top}$
represents the commanded Cartesian velocity of the EE, designed by the safety controller described below. The EE motion is represented by the discrete-time first-order model
\begin{equation}
\mathbf{x}_{k+1} =\mathbf{x}_{k} + \Delta t\,\mathbf{u}_{k},
\label{eq:ee_discrete}
\end{equation}
where $\Delta t$ denotes the control time step, and the time-varying safety field $h(\mathbf{x},t)$ defines the safe set as
\begin{equation}
\mathcal{C}(t)
=
\left\{
\mathbf{x}\in\mathbb{R}^{3}
\mid
h(\mathbf{x},t)\geq 0
\right\}.
\end{equation}
The proposed method and the Re-solve method in~\cite{bena2025geometry, Gilbert2025dynamic} share the same system model and controller formulations, differing only in how the current and predicted safety fields are updated.
In the following, we consider two safety controllers, CBF-QP and MPC-CBF.

\subsubsection{CBF-QP}
For the CBF-QP controller, a proportional controller is used as the nominal controller to generate the reference Cartesian velocity $\mathbf{u}_{\mathrm{ref}} = K_p(\mathbf{x}_{g}-\mathbf{x})$,
subject to the input limits $\mathbf{u}_{\mathrm{ref}}\in\mathcal{U}$, where $\mathbf{x}_{g}$ denotes the goal position.
Using the time-varying safety field, the safe control input is obtained from
\begin{align}
\mathbf{u}^{*}
=&
\arg\min_{\mathbf{u}}
\quad 
\|\mathbf{u}-\mathbf{u}_{\mathrm{ref}}\|_2^2
\nonumber\\
\text{s.t.}\quad 
&\mathbf{u}\in\mathcal{U},
\nonumber\\
&
\nabla h(\mathbf{x},t)^{\top}\mathbf{u}
+
\frac{\partial h(\mathbf{x},t)}{\partial t}
+
\alpha h(\mathbf{x},t)
\geq0.\label{eq:exp_cbf_qp}
\end{align}
For the experiments, $K_p=10$, $\alpha=0.2$, and the Cartesian velocity is bounded by
$ -0.7 \leq u_x,u_y,u_z \leq 0.7~\mathrm{m/s}$.
The controller is executed with a control interval of $\Delta t=0.01$ s.

\subsubsection{MPC-CBF}
The MPC-CBF controller follows the general formulation in Eq.~\eqref{eq:mpc-cbf}, instantiated here with a quadratic tracking cost and a first-order discrete-time CBF constraint. The same EE model is propagated over a horizon of $N$ steps,
$\hat{\mathbf{x}}_{i+1|k} =
\hat{\mathbf{x}}_{i|k}
+
\Delta t_{\mathrm{MPC}} \hat{\mathbf{u}}_{i|k}$.
At each control step, a sequence of future safety fields
$\{ \hat{h}_{k|k},
\hat{h}_{k+1|k},
\dots,
\hat{h}_{k+N-1|k} \}$
is generated according to the predicted motion of the moving obstacle. The MPC problem is then formulated as
\begin{align}
\min_{\{\hat{\mathbf{u}}_{i|k}\}_{i=0}^{N-1}}
 &
\sum_{i=0}^{N-1}
\left[
(\hat{\mathbf{x}}_{i|k}-\mathbf{x}_{g})^{\top}
Q
(\hat{\mathbf{x}}_{i|k}-\mathbf{x}_{g})
+
\hat{\mathbf{u}}_{i|k}^{\top}
R
\hat{\mathbf{u}}_{i|k}
\right]
\nonumber\\
&+
(\hat{\mathbf{x}}_{N|k}-\mathbf{x}_{g})^{\top}
P
(\hat{\mathbf{x}}_{N|k}-\mathbf{x}_{g})
\nonumber\\
\text{s.t.}\quad
&
\hat{\mathbf{x}}_{0|k}
=
\mathbf{x}_{k},
\nonumber\\
&
\hat{\mathbf{x}}_{i+1|k}
=
\hat{\mathbf{x}}_{i|k}
+
\Delta t_{\mathrm{MPC}}
\hat{\mathbf{u}}_{i|k},
\nonumber\\
&
\hat{\mathbf{u}}_{i|k}\in\mathcal{U},
\qquad i=0,\ldots,N-1.
\nonumber\\
&
\nabla \hat{h}_{k+i|k}
(\hat{\mathbf{x}}_{i|k})^{\top}
\hat{\mathbf{u}}_{i|k}
+
\frac{\partial \hat{h}_{k+i|k}}
{\partial t}
(\hat{\mathbf{x}}_{i|k})
\nonumber\\
&+
\alpha
\hat{h}_{k+i|k}
(\hat{\mathbf{x}}_{i|k})
\geq0.
\label{eq:exp_mpc_cbf}
\end{align}
The prediction horizon is set to $N=5$ with $\Delta t_{\mathrm{MPC}}=0.05$ s. The weighting matrices are $Q=10I_3$, $R=0.1I_3$, $P=50I_3$, with $\alpha=0.3$ and $-0.6 \leq u_x,u_y,u_z \leq 0.6~\mathrm{m/s}$.

\subsection{Experiments}

\subsubsection{Setup}
The experiments are conducted on a Universal Robots UR5e manipulator in a real-world workspace. The workspace geometry is captured using a StereoLabs Zed 2i Stereo camera positioned at $\mathbf{p}_{\mathrm{cam}} = [1.32589, -0.414097, 0.513451]$ m and represented as a 3D point cloud, from which the static safety field is constructed. The same discretisation and solver settings as described in Section~\ref{sec:simulation-safety} are used.
A rigid moving obstacle is introduced into the workspace. For visualisation purposes, the obstacle is represented in RViz as a sphere with a radius of $0.045$ m. The obstacle's position and velocity are estimated in real time using DisFlow~\cite{wu2026disflow}.
The safety field is updated asynchronously using the latest available obstacle information, and the resulting field, its spatial gradient, and time derivative are provided to the controller. The controller then computes the safe Cartesian velocity command for the UR5e EE. 

\subsubsection{Results}
\begin{figure}[t]
    \centering
    \rule{0pt}{6pt}\par
    \begin{subfigure}[b]{0.48\columnwidth}
        \centering
        \includegraphics[width=\linewidth]{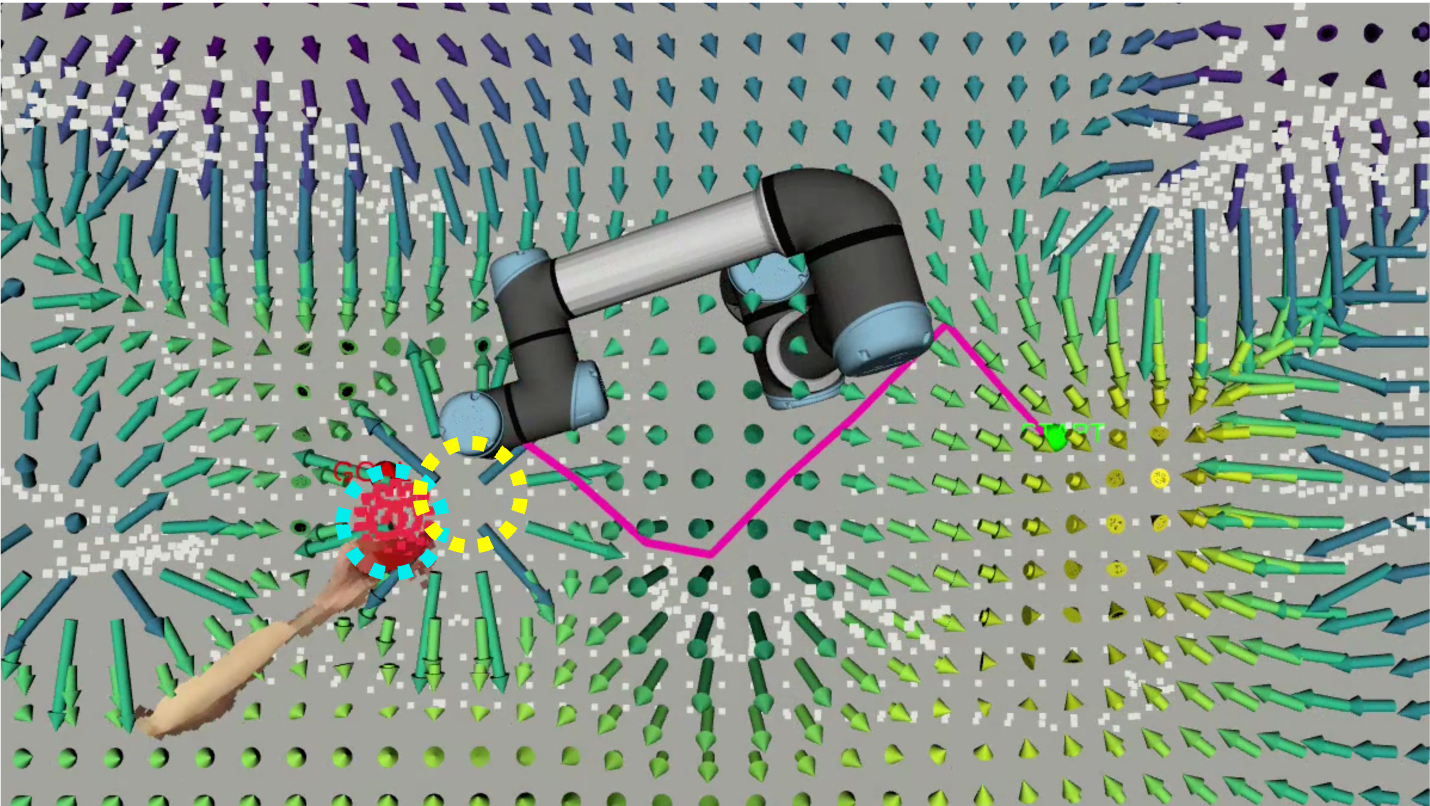}
        \caption{}
    \end{subfigure}
    \hfill
    \begin{subfigure}[b]{0.48\columnwidth}
        \centering
        \includegraphics[width=\linewidth]{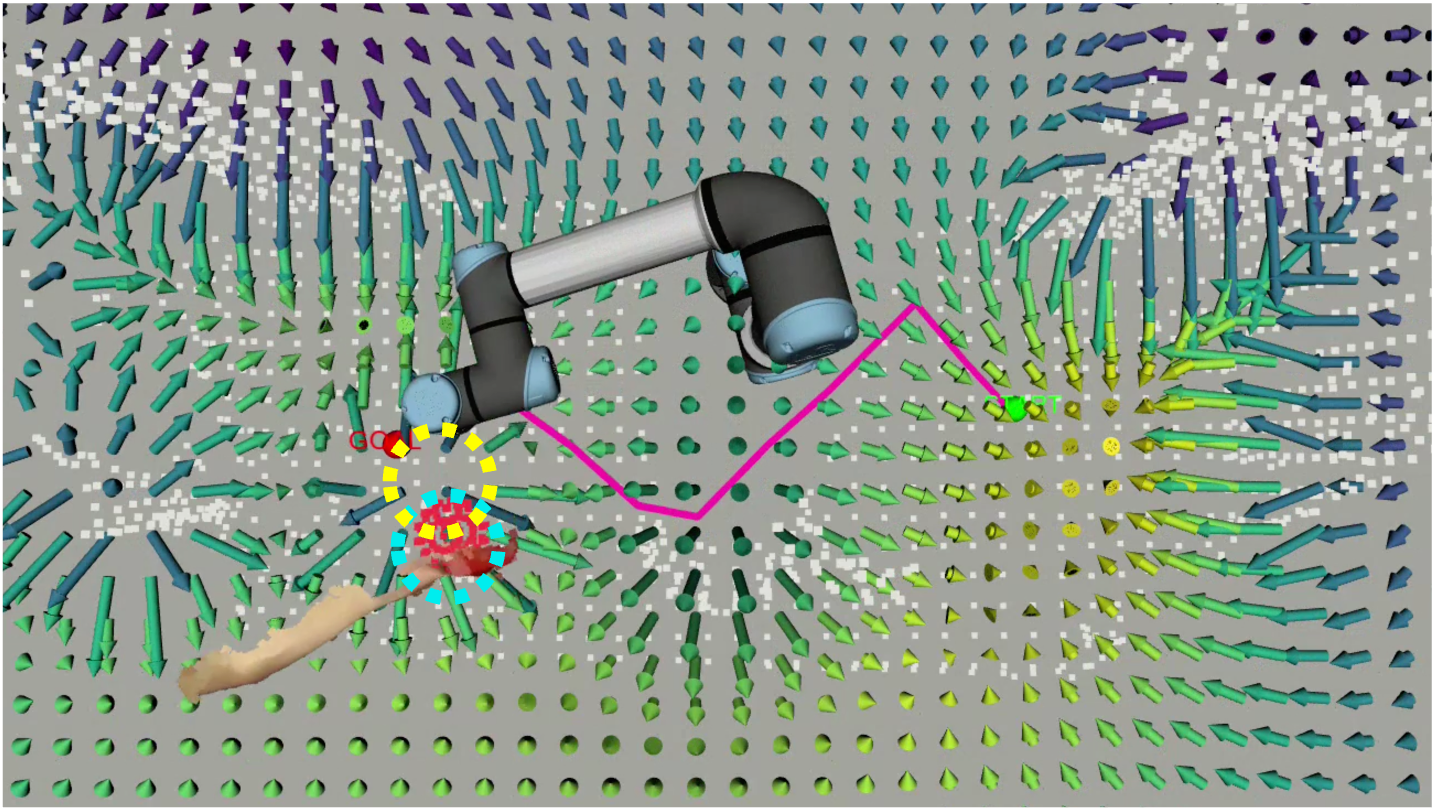}
        \caption{}
    \end{subfigure}

    \vspace{2mm}

    \begin{subfigure}[b]{0.48\columnwidth}
        \centering
        \includegraphics[width=\linewidth]{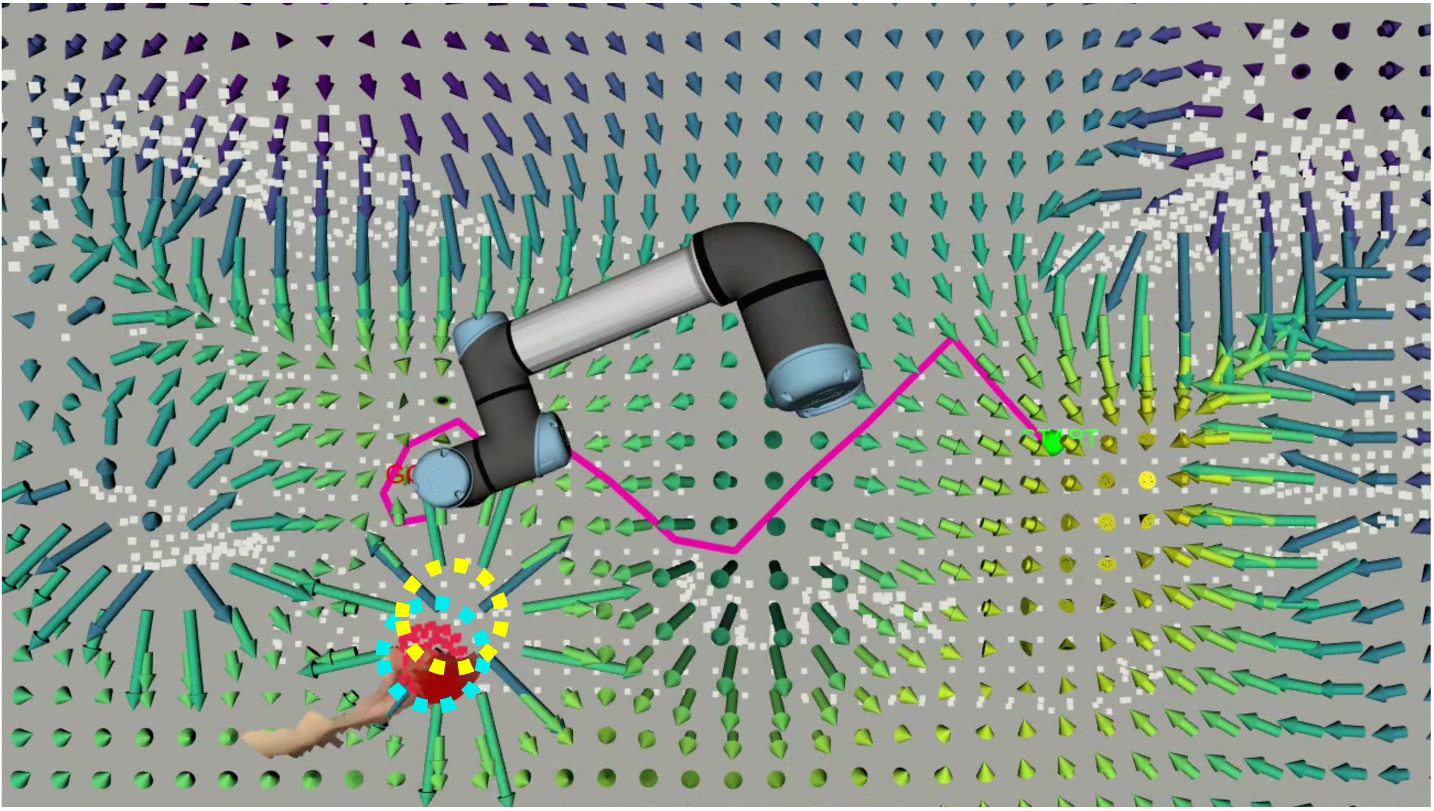}
        \caption{}
    \end{subfigure}
    \hfill
    \begin{subfigure}[b]{0.48\columnwidth}
        \centering
        \includegraphics[width=\linewidth]{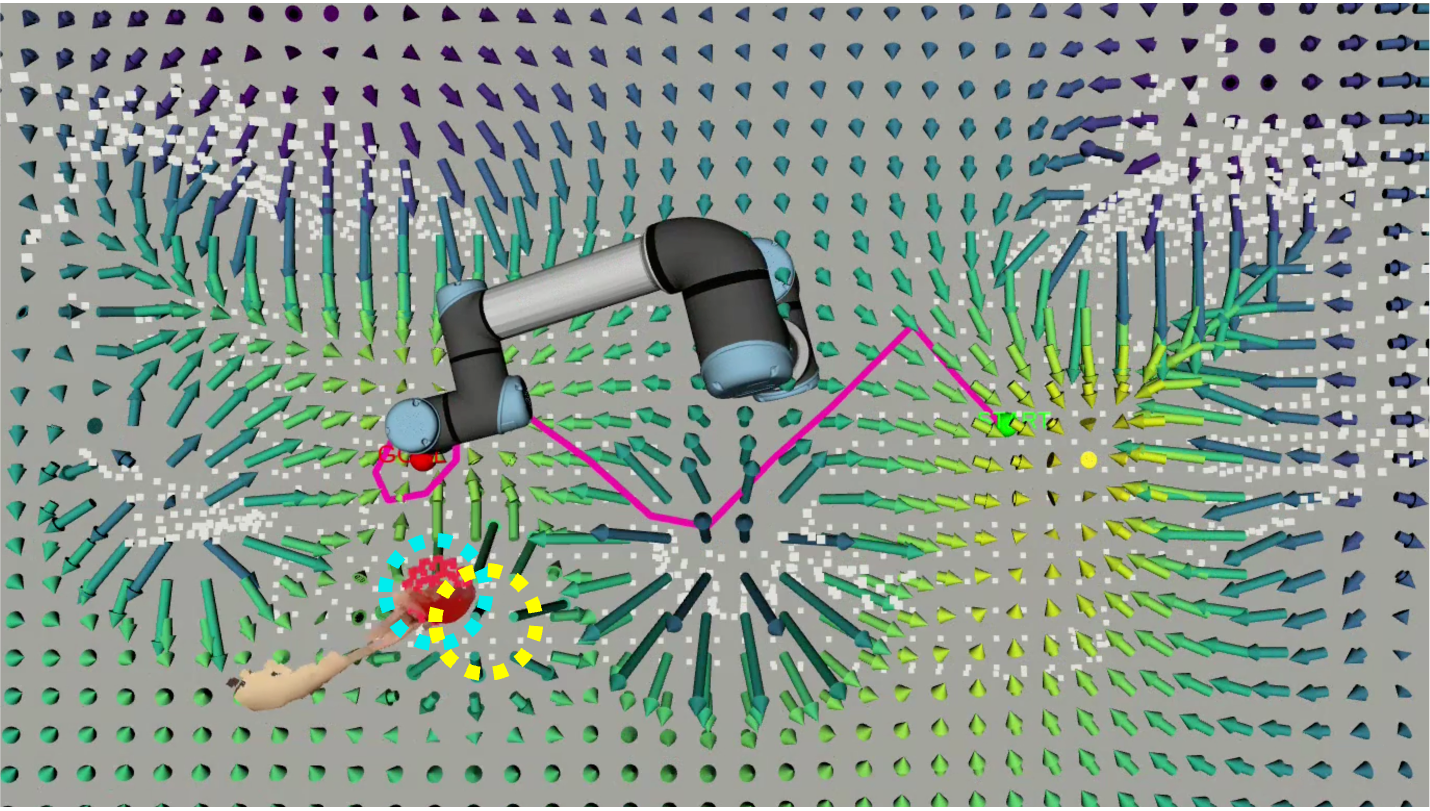}
        \caption{}
    \end{subfigure}

    \caption{Experimental results for Case 1 using MPC-CBF with the Re-solve method.}
    \label{fig:case1_resolve_mpc}
\end{figure}

Table~\ref{tab:real_world_comparison} reports the update time and task completion results for both controllers on the real UR5e platform. Consistent with the simulation results, the proposed method achieves substantially lower average and maximum update times than the Re-solve method across both CBF-QP and MPC-CBF, with average update times reduced by roughly a factor of 6--8 and maximum update times reduced by a similar margin. In Case 1, both methods successfully reach the goal under both controllers. In Case 2, however, the Re-solve method combined with MPC-CBF fails to reach the goal, while all other combinations succeed. This failure is attributed to the slower update rate of the Re-solve method, which results in outdated obstacle predictions being used within the MPC horizon, leading to overly conservative or infeasible safety constraints during fast obstacle motion. These results highlight that the computational efficiency of the safety-field update is not only beneficial for reducing computation time, but can also be critical for maintaining feasibility and task success under real-time constraints.
Fig.~\ref{fig:case1_resolve_mpc} further illustrates this effect for the Re-solve method with MPC-CBF in Case 1. The zero-level set of the safety field (yellow) consistently lags behind the moving obstacle boundary (cyan), indicating that the safety field cannot track the obstacle motion in time. Consequently, the MPC may use outdated safety information even after the obstacle has moved away.
Overall, these real-world experiments demonstrate that the efficiency gained from the proposed transport-driven update is not merely a computational advantage, but is also critical for reliable task execution when the environment changes rapidly. 

\begin{table}[htbp]
\centering
\caption{Real-world comparison of safety-field update methods.}
\label{tab:real_world_comparison}
\resizebox{\columnwidth}{!}{
\begin{tabular}{lcccc}
\toprule
& \multicolumn{2}{c}{\textbf{Ours}}
& \multicolumn{2}{c}{\textbf{Re-solve in~\cite{bena2025geometry, Gilbert2025dynamic}}} \\
\cmidrule(lr){2-3}\cmidrule(lr){4-5}
& CBF-QP & MPC-CBF & CBF-QP & MPC-CBF \\
\midrule

\multicolumn{5}{l}{\textbf{Case 1}} \\
Avg. update time (s)
& \textbf{0.0337} & \textbf{0.0366}
& 0.2134 & 0.2333 \\
Max. update time (s)
& \textbf{0.0607} & \textbf{0.0694}
& 0.2777 & 0.3588 \\
Goal reached
& \checkmark & \checkmark
& \checkmark & \checkmark \\

\midrule

\multicolumn{5}{l}{\textbf{Case 2}} \\
Avg. update time (s)
& \textbf{0.0304} & \textbf{0.0290}
& 0.2227 & 0.2229 \\
Max. update time (s)
& \textbf{0.0636} & \textbf{0.0620}
& 0.3047 & 0.2836 \\
Goal reached
& \checkmark & \checkmark
& \checkmark & $\times$ \\

\bottomrule
\end{tabular}
}
\end{table}

\section{Conclusion}\label{sec:conclusion}
This paper presented a transport-correction framework for efficiently updating PDE-based safety fields in dynamic environments. 
The proposed method avoids repeated full reconstruction of the safety field while preserving the properties required for CBF-based safety control, including a valid safe set and useful gradient information. 
The resulting time-varying safety field was further integrated into an MPC framework with DCBF constraints, allowing evolving safety constraints to be considered over the prediction horizon. 
Simulation and hardware experiments demonstrated improved computational efficiency compared with repeated PDE reconstruction and validated the effectiveness of the proposed framework for predictive safe control in dynamic environments. 
Future work will consider more general obstacle motions, uncertainty in obstacle prediction, and extensions to more complex robotic systems.

\addtolength{\textheight}{-12cm}   









\bibliographystyle{IEEEtran}
\bibliography{ref}

\end{document}